\documentclass[research]{letinbiom} 

\DateReceived{February 3, 2022}\DateAccepted{August 3, 2022}
\VolumeNo{8}\IssueNo{1}\Year{2022}

\usepackage{graphicx}
\usepackage{float}
\usepackage{listings}
\usepackage{caption}
\usepackage{subcaption}
\usepackage{siunitx}
\usepackage{booktabs}
\usepackage{makecell}
\usepackage{diagbox}
\usepackage{rotating}
\usepackage{geometry}
\usepackage{amsmath}

\usepackage[backend=biber,style=numeric, sorting=none]{biblatex}

\title{Data-driven approaches for predicting spread of infectious diseases through DINNs: Disease Informed Neural Networks}

\author{%
  Sagi \ Shaier\rlap,\inst{a}
  Maziar \ Raissi\rlap,\inst{a}
  Padmanabhan \ Seshaiyer\inst{b}
}

\institution{
  \inst{a}University of Colorado, Boulder, USA;
  \inst{b}George Mason University, Fairfax, Virginia, USA
}

\correspondence{Sagi Shaier}{Sagi.Shaier@Colorado.edu}

\authorheading{S.~Shaier, M.~Raissi, P.~Seshaiyer}


\abstract{
In this work, we present an approach called Disease Informed Neural Networks (DINNs) that can be employed to effectively predict the spread of infectious diseases. 
We build on the application of physics informed neural network (PINNs) to SIR compartmental models and expand it to a scaffolded family of mathematical models describing various infectious diseases. We show how the neural networks are capable of learning how diseases spread, forecasting their progression, and finding their unique parameters (e.g. death rate). To demonstrate the robustness and efficacy of DINNs, we apply the approach to eleven highly infectious diseases that have been modeled in increasing levels of complexity. Our computational experiments suggest that DINNs is a reliable candidate to effectively learn the dynamics of their spread and forecast their progression into the future from available real-world data. Code and data can be found here: \url{https://github.com/Shaier/DINN}
}

\begin{document}

\maketitle


\section{Introduction}

Understanding the early transmission dynamics of infection diseases has never been more important in history as of today. The outbreak of severe acute respiratory syndrome coronavirus 2 (SARS-CoV-2) that led to several million confirmed cases across the globe has challenged us to re-envision how we model, analysis and simulate infectious diseases and evaluate the effectiveness of non-pharmaceutical control measures as important mechanisms for assessing the potential for sustained transmission to occur in new areas.

 An important contribution to the mathematical theory of epidemics was developed by Kermack–McKendrick epidemic model of 1927  \cite{kermack1927contribution}. This was considered one of the earliest attempts to formulate a simple mathematical model to predict the spread of an infectious disease where the population being studies is divided into compartments namely a susceptible class $S$, an infective class $I$, and a removed class $R$. 

\begin{figure}[h!]
\begin{center}
  \includegraphics[scale=0.3]{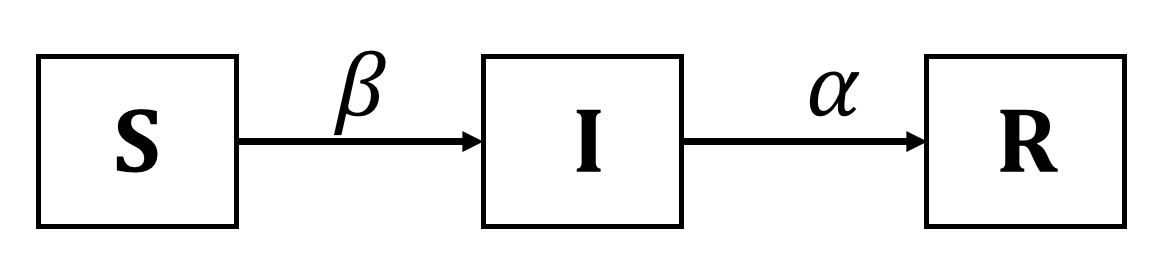}
  \caption{Compartmental Model for SIR model}\label{figSIR}
  \end{center}
\end{figure}

 This simple SIR epidemic model can be illustrated in compartments as in Figure \ref{figSIR}. Not only was it capable of generating realistic single-epidemic outbreaks but also provided important theoretical epidemiological insights.
In Figure \ref{figSIR}, it is assumed that each class resides within exactly one compartment and can move from one compartment to another. The dynamics of the three sub-populations $S(t)$, $I(t)$ and $R(t)$ may be described by the following SIR model given by first order coupled ordinary differential equations (ODE) \cite{hethcote2009basic, brauer2012mathematical, martcheva2015introduction, brauer2017mathematical}:
\begin{eqnarray}\label{SIR}
\frac{dS}{dt} &=& -\beta \; S \; I \nonumber \\
\frac{dI}{dt} &=& \beta \; S \; I - \alpha \; I\\
\frac{dR}{dt} &=& \alpha \; I \nonumber
\end{eqnarray}
Note that this closed system does not allow any births/deaths. This SIR model in system (\ref{SIR}) is fully specified by prescribing the {\em transmission} rate $\beta$ and {\em recovery} rate $\alpha$ along with a set of initial conditions $S(0)$, $I(0)$ and $R(0)$.  The total population $N$ at time $t=0$ is given by $N = S(0) + I(0) + R(0)$. Adding all the equations in system (\ref{SIR}), we notice that $N^{\prime}(t) = 0$
and therefore $N(t)$ is a constant and equal to its initial value. One can further assume $R(0) = 0$ since no one has yet had a chance to recover or die. Thus a choice of $I(0)=I_0$ is enough to define the system at $t=0$ since then $S_0 = N - I_0$. 

Following the influenza pandemic, several countries and leading organizations increased funding and attention to finding cures for infectious diseases in the form of vaccines and medicines. Along with these policy implementations, newer modified SIR models for mathematical epidemiology continued to evolve, particularly for those diseases that are categorized as re-emerging infections \cite{castillo2002mathematical}, those that are spread through sexual transmission such as HIV \cite{castillo2013mathematical, luo2016bifurcations}, those that are spread through vectors such as mosquitoes such as Malaria or Dengue \cite{li2011malaria, chowell2007estimation}, those that can spread through both sexual and vector transmissions such as Zika \cite{padmanabhan2017mathematical, padmanabhan2017computational}, and those that can be spread by viruses, including SARS and MERS \cite{dye2003modeling, alshakhoury2017mathematical}. Diseases were also categorized according to the rate at which they spread, for example, super-spreader diseases. This point is especially relevant to COVID-19 \cite{ohajunwa2020mathematical,ohajunwa2021mathematical}, categorized as a super-spreader based on the disproportionately fast rate and large (and growing) number of infected persons. 

Along with the development of mathematical modeling, there have been a variety of approaches that have been introduced to estimate the parameters such as the transmission, infection, quarantine and recovery using real-data. These include nonparametric estimation \cite{smirnova2019forecasting}, optimal control \cite{neilan2010introduction}, Bayesian frameworks \cite{coelho2011bayesian,akman2016examination}, partical swarm optimization \cite{ akman2018parameter}, inverse methods, least-squares approach, agent-based modeling, using final size calculations \cite{bell1990mathematical, pollicott2012extracting, yong2015agent, martcheva2015introduction}. Also, researchers have employed a variety of statistical approaches including maximum-likelihood, Bayesian inference and Poisson regression methods \cite{huang2006hierarchical, longini1988statistical, hadeler2011parameter, o2014fitting, capaldi2012parameter,}. Some of this work also showed that the precision of the estimate increased with the number of outbreaks used for estimation \cite{o2014fitting}. To determine the relative importance of model parameters to disease transmission and prevalence, there has also been work around {\em sensitivity analysis} of the parameters using techniques such as Latin Hypercube Sampling and Partial Rank Correlation Coefficients analysis with the associated mathematical models \cite{blower1994sensitivity, mckay2000comparison, chitnis2008determining,}.  While there have been significant advances in estimating parameters, there is still a great need to develop efficient, reliable and fast computational techniques. 


The dominant algorithm associated with the advancements in artificial intelligence ranging from computer vision \cite{goodfellow2014generative, NIPS2012_c399862d, redmon2016look, tan2020efficientdet} to natural language processing \cite{devlin2019bert, vaswani2017attention} has been the neural networks (NN).  A main reason for it is its behavior as a universal function approximator \cite{HORNIK1989359}.  However, this field is largely relying on huge amounts of data and computational resources. Recent approaches \cite{RAISSI2019686} have been shown to be successful in combining the best of both fields. That is, using neural networks to model nonlinear systems, but reducing the required data and by constraining the model's search space with known knowledge such as a system of differential equations. 

Along with this, there have also been several works recently showing how differential equations can be learned from data. For example, \cite{osti_1333570} used a deep neural network to model the Reynolds stress anisotropy tensor, \cite{E_2017} solved for parabolic PDEs and backward stochastic differential equations using reinforcement learning, and \cite{hagge2017solving} solved ODEs using a recurrent neural network.  Additionally, \cite{Raissi_2018, RAISSI2019686} developed physics informed models and used neural networks to estimate the solutions of such equations. Using this, recently such physics informed neural network approaches were applied for the first time to estimating parameters accurately for SIR model applied to a benchmark application \cite{raissi2019parameter}. The  Physics Informed Neural Network approaches have also been recently used for studying the dynamics of COVID-19 involving human-human and human-pathogen interaction \cite{NguyenRaissiSeshaiyer}.

Building on this, a unified approach called DINNs: Disease Informed Neural Networks is introduced in this work and systematically applied to some increasingly complex governing system of differential equations describing various prominent infectious diseases over the last hundred years. These systems vary in their complexity, ranging from a system of three to nine coupled equations and from a few parameters to over a dozen. For illustration of the application of DINNs, we introduce its application to COVID, Anthrax, HIV, Zika, Smallpox, Tuberculosis, Pneumonia, Ebola, Dengue, Polio, and Measles. Our contribution in this work is three fold. First, we extend the recent physics informed neural networks (PINNs) approach to a large family of infectious diseases. Second, we perform an extensive analysis of the capabilities and shortcomings of DINNs on diseases. Lastly, we show the ease at which one can use DINNs to effectively learn the dynamics of the disease and forecast its progression a month into the future from real-life data.

The paper is structured as follows. In Section \ref{sec:Background} we review the necessary background information. Section \ref{sec:dinn} introduces DINNs and presents our technical approach. Section \ref{Experiments} shows the application of DINNS to some of the benchmark models through computational experiments. Lastly, we conclude with a summary in Section \ref{sec:conclusion}.
\section{Background Models and Methods}
\label{sec:Background}
A grand challenge in mathematical biology and epidemiology, with great opportunities for researchers working on infectious disease modeling, is to develop a coherent framework that enables them to blend differential equations such as the system (\ref{SIR}) with the vast data sets now available. 

Noting that $\displaystyle\frac{dS}{dt} < 0$ for all $t$, the susceptible population $S(t)$ is monotonically decreasing and always declining
independently of the initial condition S(0). Also, we have $\displaystyle\lim_{t \rightarrow \infty} S(t) = S_{\infty}$. This quantity is refered to as {\it final size} of the epidemic \cite{brauer2012mathematical}. Also, when $S = \displaystyle\frac{\alpha}{\beta}$ the second equation in system (\ref{SIR}), $\displaystyle\frac{dI}{dt} = 0$ which indicates that $I(t)$ has a stationary point at some maximum time. On the other hand, the number of infected individuals may be monotonically decreasing to zero, or may
have non-monotone behavior by first increasing to some maximum level, and then decreasing to zero. One may
note that the spread starts to increase if $\displaystyle\frac{dI}{dt} > 0$. This yields the following necessary and sufficient  condition
for an initial increase in the number of infectives given by $\displaystyle\frac{\beta S(0)}{\alpha} > 1$. Hence if $S_0 < \displaystyle\frac{\alpha}{\beta}$, the infection dies out and there is no epidemic. The last equation in system (\ref{SIR}) also indicates that the recovered individuals also have monotone behavior, independent of $R(0) = R_0$. Since $\displaystyle\frac{dR}{dt} > 0$ for all $t$, the number of recovered is always increasing monotonically. Since we know that this number is constrained by the total population $N$, we also have $\displaystyle\lim_{t \rightarrow \infty} R(t) = R_{\infty}$. Since the total population $N = S_{\infty} + R_{\infty} = S_0 + R_0$, one can derive \cite{brauer2012mathematical}
\begin{equation}\label{eqsinfty}
S_{\infty} = S_0 e^{-\frac{\beta}{\alpha} \; (S_0 + I_0 - S_{\infty})}
\end{equation}
as well as the the maximum number of infected individuals $I_{max}$ reached in the epidemic which occurs at $S = \frac{\alpha}{\beta}$:
\begin{equation}\label{eqImax}
I_{max}= -\frac{\alpha}{\beta} + \frac{\alpha}{\beta}  \ln \left(\frac{\alpha}{\beta} \right) + I_0 + S_0 - \frac{\alpha}{\beta}  \ln S_0
\end{equation}
\subsection{Approaches for estimating rates}
There are multiple approaches that can be used to estimate parameters $\alpha$ and $\beta$ in system (\ref{SIR}). Assuming the epidemic was initiated by one infected individual infecting $n$ other individuals a day later, a crude approximation could be to use $\displaystyle\frac{dS}{dt} \approx -n$ per individual per day. Given $S_0$ and $I_0$, one can then estimate the initial transmission rate to be 
\[
\beta = -\displaystyle\frac{\displaystyle\frac{dS}{dt}}{S_0 I_0} = \displaystyle\frac{n}{S_0 I_0}
\]
If the infected individuals are isolated within $d$ days of becoming sick, one may estimate that $\displaystyle\frac{1}{d}$ of the infected population was removed each day, or $\alpha = \displaystyle\frac{1}{d}$ per day. This then yields the ratio of $\displaystyle\frac{\alpha}{\beta}$. Using these values one can then plot the dynamics of the model predictions compared to the data using the system (\ref{SIR}).  Moreover one can use equations (\ref{eqsinfty}) and (\ref{eqImax}) to determine $S_{\infty}$ and $I_{max}$. 

Another approach to determine the rates that can be employed is by noticing the population that seem to have escaped the epidemic which could serve as the $S_{\infty}$. 
Then, using equation (\ref{eqsinfty}) with the given dataset one can determine the ratio:
\begin{equation}\label{eqbetaalpha}
\frac{\beta}{\alpha} = \frac{\ln\left(\frac{S_0}{S_{\infty}}\right)}{N - S_{\infty}} 
\end{equation}
Then assuming as before that the infected were quarantined for about $d$ days as infectious individuals, one can find the recovery rate to be $\alpha = \displaystyle\frac{1}{d}$ which can then used to estimate $\beta$ using (\ref{eqbetaalpha}).

It must be pointed out that this data set consists of a closed population. It must be also noted that all these models assumed that the recovery rate $\alpha$ can be computed heuristically. However, there are also methods in the literature that can help to estimate the parameters including $S_0$, $\alpha$ and $\beta$ in system (\ref{SIR}) by minimizing the deviation between the SIR model out and a given data set. One such software implementation is Berkeley Madonna \cite{macey2000} which has been shown to fit the data using the fitting parameter as $\alpha$ \cite{yonganuj2015}.

Yet another approach for parameter estimation is an optimization algorithm that employs a least-squares minimization approach to estimate optimal parameters. Specifically, one can employ an unconstrained nonlinear optimization algorithm such as the Nelder-Mead algorithm which searches for a local minimum using a regression approach. This direct search method attempts to minimize a function of real variables using only function evaluations without any derivatives \cite{Nsoesie2013}. The minimized objective function is represented by differences in the daily infected counts from observed data and the computer simulated data. 

Clearly, from these estimation approaches outlined so far, there can be variations in the ability of the dynamics of the computed values of infected population to track the true data for the various combination of parameters. As noted, these parameters are often calculated through heuristic methods in some of these algorithms and may not be optimal. Also, all the methods that have been discussed so far assumed the prior knowledge of initial number of each of the human sub-populations including the Susceptible $S_0$, Infected $I_0$ and Recovered $R_0$. 

\begin{figure}[h!]
\begin{center}
  \includegraphics[scale=0.75]{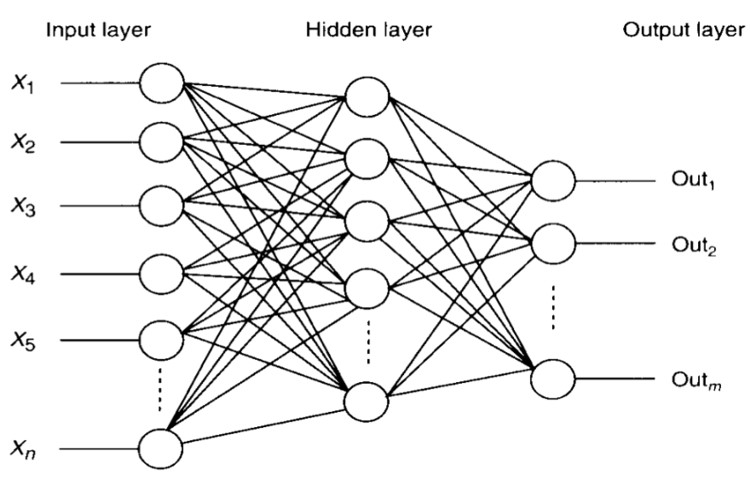}
  \vspace{-0.5cm}
  \caption{An illustration of a neural network}\label{nn}
  \end{center}
\end{figure}

In this work, we introduce deep learning as an alternative and powerful approach, that employs {\em neural networks} which is a system of decisions modeled after the human brain {\cite{lecun2015deep}}. Consider the illustration shown in Figure \ref{nn}.
The first layer of perceptrons first weigh and bias the input which can be observed values of infected data. The next layer then will make more complex decisions based off those inputs, until the final decision layer is reached which generates the outputs which can correspond to the values of parameters such as $\beta$ and $\alpha$. In this research, we implement a physics informed neural network based approach which makes decisions based on appropriate activation functions depending on the computed {\em bias} ($b$) and {\em weights} ($w$). The network then seeks to minimize the mean squared error of the regression with respect to the weights and biases by utilizing gradient descent type methods used in conjunction with software such as tensorflow. While there is currently a lot of enthusiasm about “big data”, useful data in infectious diseases is usually “small” and expensive to acquire. In this work, we will describe how one can apply such physics informed neural network based deep learning approaches specifically to infectious diseases using DINNs and apply it to a real-world example to estimate optimal parameters, namely the transmission and recovery rates, in the SIR model.

\section{Disease Informed Neural Networks}
\label{sec:dinn}
\begin{figure}[htbp]
    \centering
    \includegraphics[scale=0.5]{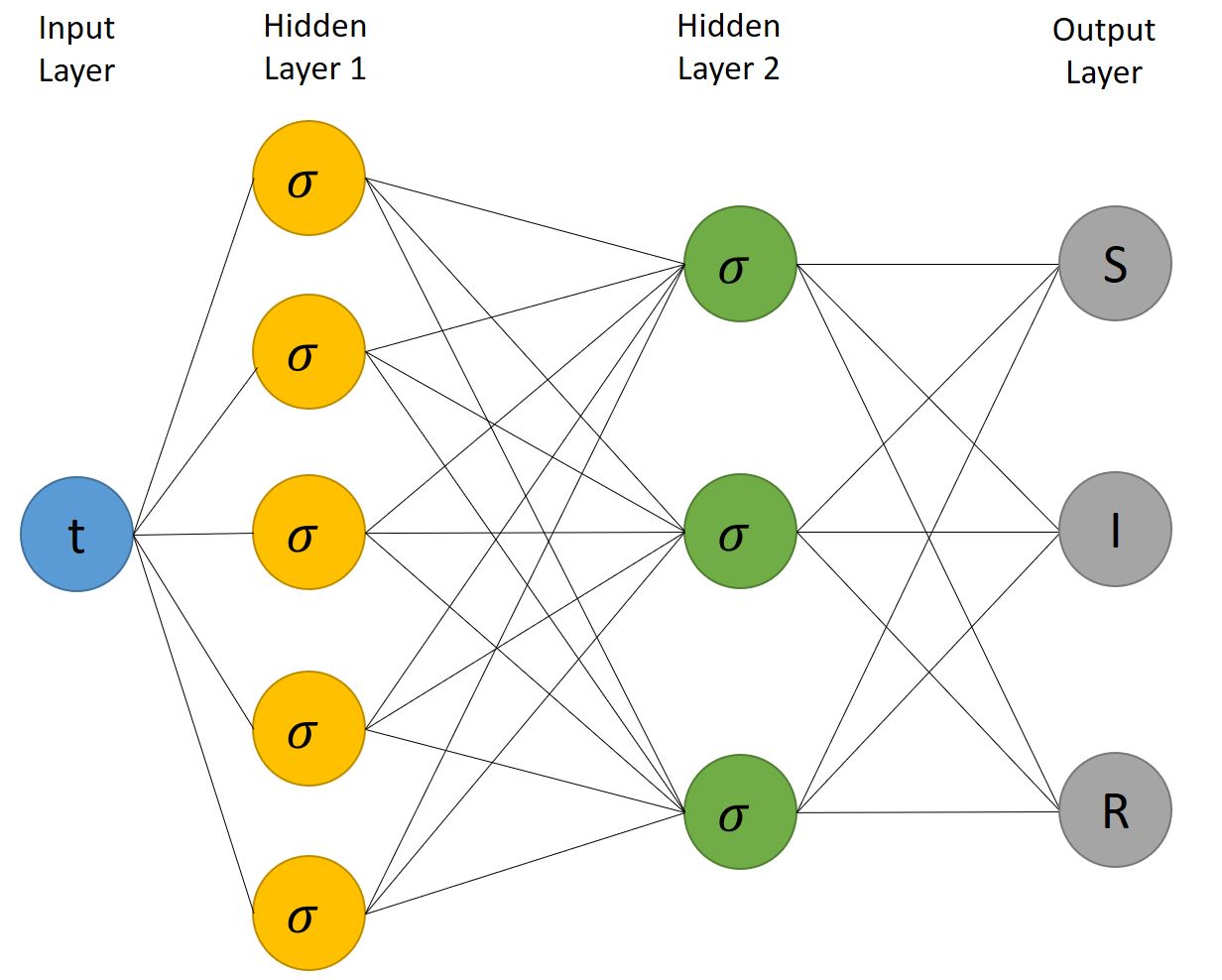}
    \caption{A Simple DINNs Architecture (input --- time, has a size of 1 and the output can vary in size (S,I,R)).}
    \label{Deep Neural Net}
\end{figure}

In this section, we present the DINNs methodology (sample architecture can be seen in figure \ref{Deep Neural Net}). Subsection \ref{Neural Networks subsection} briefly discusses background information for neural networks. Subsection \ref{DINNs approach} provides an overview of the DINNs approach and outlines the algorithm, associated loss functions, and training information. 

\subsection{Neural Networks Architechture}
\label{Neural Networks subsection}
Briefly speaking, neural network is an attempt to mimic the way the human brain operates. The general fully connected model is organized into layers of nodes (i.e. neurons) where each node in a single layer is connected to every node in the following layer (except for the output layer), and each connection has a particular weight. The idea is that deeper layers capture richer structures \cite{eldan2016power}. A neuron takes the sum of weighted inputs from each incoming connection (plus a bias term), applies an activation function (i.e nonlinearity), and passes the output to all the neurons in the next layer. Mathematically, each neuron's output looks as follows
\begin{gather*}
\text{output} = \sigma\left(\sum_{i=1}^{n}x_i w_i + b \right)
\end{gather*}
where $n$ represents the number of incoming connections, $x_i$ the value of each incoming neuron, $w_i$ the weight on each connection, $b$ is a bias term, and $\sigma$ is referred to as the activation function. 

A schematic representation of the resulting \emph{disease informed neural networks} is given in Figure \ref{Deep Neural Net}. Note that for simplicity of illustration figure \ref{Deep Neural Net} depicts a network that comprises of 2 hidden layers with 5 neurons in the first hidden layer and 3 in the second. Networks with this kind of many-layer structure - two or more hidden layers - are called {\em deep neural networks}. These neurons in the network may be thought of as holding numbers that are calculated by a special {\em activation function} that depends on suitable {\em weights} and {\em biases} corresponding to each connection between neurons in each layer. With prior knowledge of such an activation function, the problem boils down to identifying the weights and biases that correspond to computed values of infected data that is close to the observed values. The three sub-populations are approximated by the deep neural network with calculus on computation graphs using a backpropogation algorithm \cite{hecht1992theory, schmidhuber2015deep, goodfellow2016deep}. 

Inspired by recent developments in \emph{physics-informed deep learning} \cite{Raissi_2018, RAISSI2019686}, we propose to leverage the hidden physics of infectious diseases \eqref{SIR} and infer the latent quantities of interest (i.e., $S$, $I$, and $R$) by approximating them using deep neural networks. This choice is motivated by modern techniques for solving forward and inverse problems associated with differential equations, where the unknown solution is approximated either by a neural network  or a Gaussian process. Following these approaches, we approximate the latent function
\[
t \longmapsto (S,I,R)
\]
by a deep neural network and obtain the following DINNs corresponding to equation \eqref{SIR} and the total population $N = S + I + R$, i.e.,
\begin{equation}\label{eq:NN}
\arraycolsep=1.4pt\def\arraystretch{1.5}
\begin{array}{l}
E_1 := \frac{dS}{dt} + \beta \; S \; I, \\
E_2 := \frac{dI}{dt} - \beta \; S \; I + \alpha \; I,\\
E_3 := \frac{dR}{dt} - \alpha \; I,
\end{array}
\end{equation}

We acquire the required derivatives to compute the residual networks $E_1$, $E_2$ and $E_3$ (disease informed) by applying the chain rule for differentiating compositions of functions using automatic differentiation \cite{baydin2018automatic}.  In our computations, we employed
a densely connected neural network, which takes the input variable $t$ and outputs $S, I$, and $R$. 

It is worth highlighting that parameters $\alpha$ and $\beta$ of the differential equations turn into parameters of the resulting disease informed neural networks $E_1$, $E_2$ and  $E_3$. The total loss function is composed of the regression loss corresponding to $S$, $I$ and $R$ and the loss imposed by the differential equations system (\ref{eq:NN}). Moreover, the gradients of the loss function are back-propogated through the entire network to train the parameters using a gradient-based optimization algorithm. As will be explained next, we will assume that the only observables are noisy data that we will use in conjunction with the neural networks for S, I, R to estimate parameters $\alpha$, $\beta$ and $\gamma$ by minimizing the sum of squared errors loss function. The idea employed builds on Physics Informed Neural Networks that can embed the knowledge of any physical law that govern a given data-set in the learning process \cite{Raissi_2018, RAISSI2019686}.

\subsection{DINNs for Parameter Estimation}
\label{DINNs approach}
The predictive capability of any algorithm is measured partially by its robustness to unknown data. A dataset for known parameters can be simulated by solving a system of equations in a forward fashion and potentially adding some noise. If that is provided to any parameter estimation algorithm, the efficacy of the algorithm can be determined by how well it is able to predict the true values for a wide range of starting guesses.

For simplicity, we generated data by solving the systems of disease ODEs using LSODA algorithm \cite{lsoda}, the initial conditions, and the true parameters corresponding to each disease (e.g. death rate) from the literature. This limited dataset ($50-100$ points) corresponds to the SIR compartments. To make our neural networks disease informed, once the data was obtained we introduced it to our neural network without any prior knowledge of the transmission and recovery parameters. It is worth noting that in this formulation there are no training, validation, and test datasets, such as in most common neural networks training. Instead, the model is trained from data of how the disease is spread over time. The model then learned the system, and predicted the parameters that generated them. Since in many of these systems there exist a large set of parameters that can generate them, we restricted our parameters to be in a certain range around the true value. That is, to show that our model can in fact identify the systems and one set of parameters that match the literature they came from. However, our method is incredibly flexible in the sense that adding, modifying, or removing such restrictions can be done with one simple line of code. Additionally, we used nearly a years worth of real data aggregated over every US state and accurately predicted a month into the future of COVID transmission.
Next we employ Literate programming style that is intended to facilitate presenting parts of written code in the form of a narrative \cite{knuth1984literate}.
DINNs takes the form
\begin{lstlisting}[language=Python]
def net_sir(time_array):
    SIR = neural_network(time_array)
    return SIR

def net_f(time_array):
    dSdt = torch.grad(S, time_array)
    dIdt = torch.grad(I, time_array)
    dRdt = torch.grad(R, time_array)
    
    f1 = dSdt - (-beta SI)
    f2 = dIdt - (beta SI - alpha I)
    f3 = dRdt - (alpha I)
    return f1, f2, f3, S, I, R

\end{lstlisting}




The input of the neural network {\tt net\_{sir}} is a batch of time steps (e.g. $[0, 1, 2, ..., 100]$), and the output is a tensor (e.g. [S,I,R]) that represents what the network believes the disease's compartments look like at each time step. Here, {\tt net\_{f}} bounds the NN by forcing it to match the environment's conditions (e.g. $f_1, f_2, f_3$). These $f_i$ corresponds to the $E_i$ that was described earlier.

The parameters of the neural network {\tt net\_sir} and the network {\tt net\_f} can be learned by minimizing the mean squared error loss given by
\begin{gather*}
\text{MSE} = \text{MSE}_\text{\tt net\_sir} + \text{MSE}_\text{\tt net\_f}
\end{gather*}
where
\begin{eqnarray*}
  \text{MSE}_{\mathrm{\tt net\_sir}}
    &=& \frac{1}{N_{\mathrm{\tt net\_sir}}} 
      \biggl[ \sum_{i=1}^{N_{\mathrm{\tt net\_sir}}} \bigl\lvert \mathrm{\tt net\_sir}(\mathrm{time\_array}^i) - \mathrm{SIR}^i \bigr\rvert^2 \biggr] \\
  \text{MSE}_{\mathrm{\tt net\_f}}
    &=& \frac{1}{N_{\mathrm{\tt net\_f}}} 
      \biggl[ \sum_{i=1}^{N_{\mathrm{\tt net\_f}}} \bigl\lvert  \mathrm{\tt net\_f}^i \bigr\rvert^2 \biggr]
\end{eqnarray*}
That is, minimizing the loss

\begin{eqnarray*}
\text{loss} = \text{mean((S}_\text{actual} - \text{S}_\text{predict})^2) &+&  \text{mean((I}_\text{actual} - \text{I}_\text{predict})^2) \\
&+&
\text{mean((R}_\text{actual} - \text{R}_\text{predict})^2) \\
&+&
\text{mean((f}_1)^2) + \text{mean((f}_2)^2) + \text{mean((f}_3)^2)
\end{eqnarray*}

Here, “actual” and “predict" refer to the actual data that the model was provided with and the prediction the model computed, respectively. DINNs also leverages the automatic differentiation that neural networks are trained on to compute the partial derivatives of each S,I,R with respect to time. The neural networks themselves will consist of multiple fully connected layers with a multiple neurons each depending on the complexity of the system and {\em rectified linear activation function} (ReLU) activation in between.

\section{Computational Experiments with DINNs}
\label{Experiments}
Most mathematical models describing the spread of a disease employ classical compartments, such as the Susceptible-Infected-Recovered (SIR) or Susceptible-Exposed-Infected-Recovered (SEIR) structure described as an ordinary differential equation system \cite{brauer2001basic}. Over the past several months there have been a variety of compartmental models that have been introduced as modified SEIR models to study various aspects of COVID-19 including containment strategies \cite{maier2020effective}, social distancing \cite{matrajt2020evaluating} and the impact of non-pharmaceutical interventions and the social behavior of the population \cite{ohajunwa2020mathematical, ohajunwa2021mathematical}. Along with these there have been a lot of work on modified SIR models as well including  the SIRD model \cite{fernandez2020estimating, Anastassopoulou2019, sen2021use, chatterjee2021studying}. Next, to investigate the performance of DINNs, we apply DINNs on a simple SIRD model describing COVID-19 dynamics \cite{Anastassopoulou2019}. 

\subsection{Applying DINNS to an SIRD model applied to COVID-19}
Consider following differential equation system describing the SIRD system. Here $\alpha$ is the transmission rate, $\beta$ is the recovery rate and $\gamma$ is the death rate from the infected individuals \cite{Anastassopoulou2019}. $N$ represents the total population.
\begin{eqnarray}
\frac{dS}{dt} &=& - (\alpha / N) S I  \label{eqnSIRD-S}\\
\frac{dI}{dt}  &=& (\alpha / N) S I - \beta I - \gamma I \label{eqnSIRD-I}\\
\frac{dR}{dt}  &=& \beta I \label{eqnSIRD-R}\\
\frac{dD}{dt}  &=& \gamma I \label{eqnSIRD-D}
\end{eqnarray}

The neural networks we considered,  are fairly simple, consisting of 8 fully connected layers with either 20 or 64 neurons each depending on the complexity of the system and {\em rectified linear activation function} (ReLU) activation in between. Since the data is relatively small, our batch size contained the entire time array. The networks were trained on Intel(R) Xeon(R) CPU @ 2.30GHz, and depending on the complexity of the system the training time ranged from 30 minutes to 58 hours, which could be accelerated on GPUs and TPUs. That is, to learn both a system and its unknown parameters. However if the parameters are known, the training time to solely learn the system can be as short as 3 minutes. We used Adam optimizer \cite{kingma2014adam}, and PyTorch’s CyclicLR as our learning rate scheduler, with mode = “{\tt exp\_range}", {\tt min\_lr} ranging from $1 \times 10^{-6}$ to $1 \times 10^{-9}$ depending on the complexity of the system, {\tt max\_lr} = $1 \times 10^{-3}$, gamma=0.85, and {\tt step\_size\_up}=1000. In the next sections we will refer to “{\tt min\_lr}" simply as “learning rate". It is important to note that some diseases' systems were much more difficult for DINNs to learn (e.g. Anthrax considered later) and further training exploration such as larger/smaller learning rate, longer training, etc. may be needed to achieve better performance.

\subsubsection{Influence of ranges in Parameter Estimation}
Given that most models may include a large set of parameters, it is important to consider ranges for each of them.  Hence, we restricted our parameters to be in a certain range to show that our model can learn the set that was used in the literature. 
First, we experimented with various parameter ranges to identify the influence they had on the model. In the following we used a 4 layer neural network with 20 neurons each, $1 \times 10^{-6}$ learning rate, 100 data points, and the models were trained for 700,000 iterations (taking roughly 30 minutes). In our experiments we report two kinds of relative MSE loss errors. The first, “Error NN", is the error on the neural network's predicted system. The second, “Error learnable parameters", is the error on the system that was generated from the learnable parameters. That is, using LSODA algorithm to generate the system given the neural networks' parameters (e.g. $\alpha$).

\begin{figure}[!htb]
\centering
\includegraphics[width = 3in]{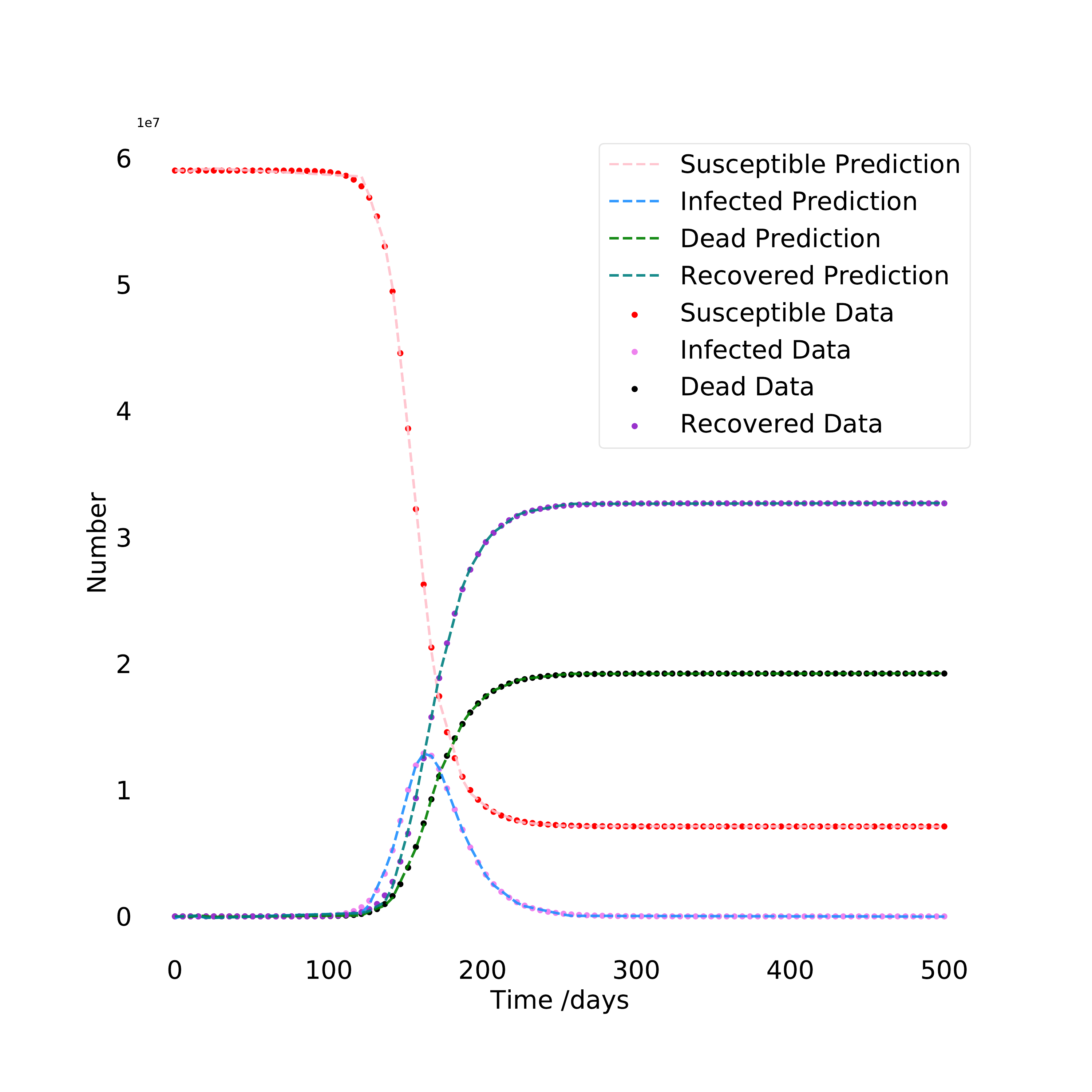} 
\includegraphics[width = 3in]{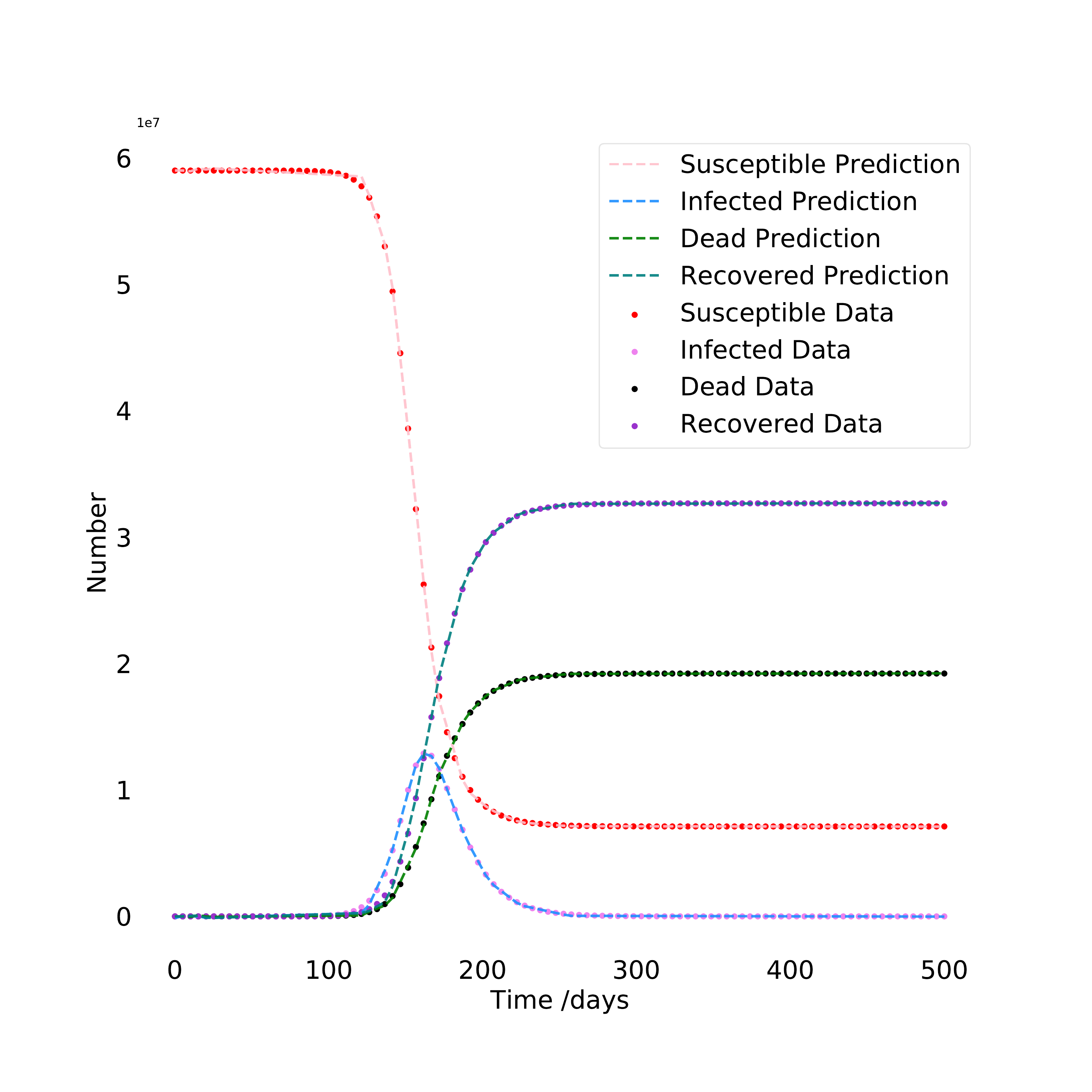}
\caption{0\% search range for NN Output (Left panel) vs LSODA generation from Learnable Parameters (Right Panel)}
\label{0 parameters ranges figure}
\end{figure}

\begin{figure}[!htb]
\centering
\includegraphics[width = 3in]{images/param_known_nn.pdf} 
\includegraphics[width = 3in]{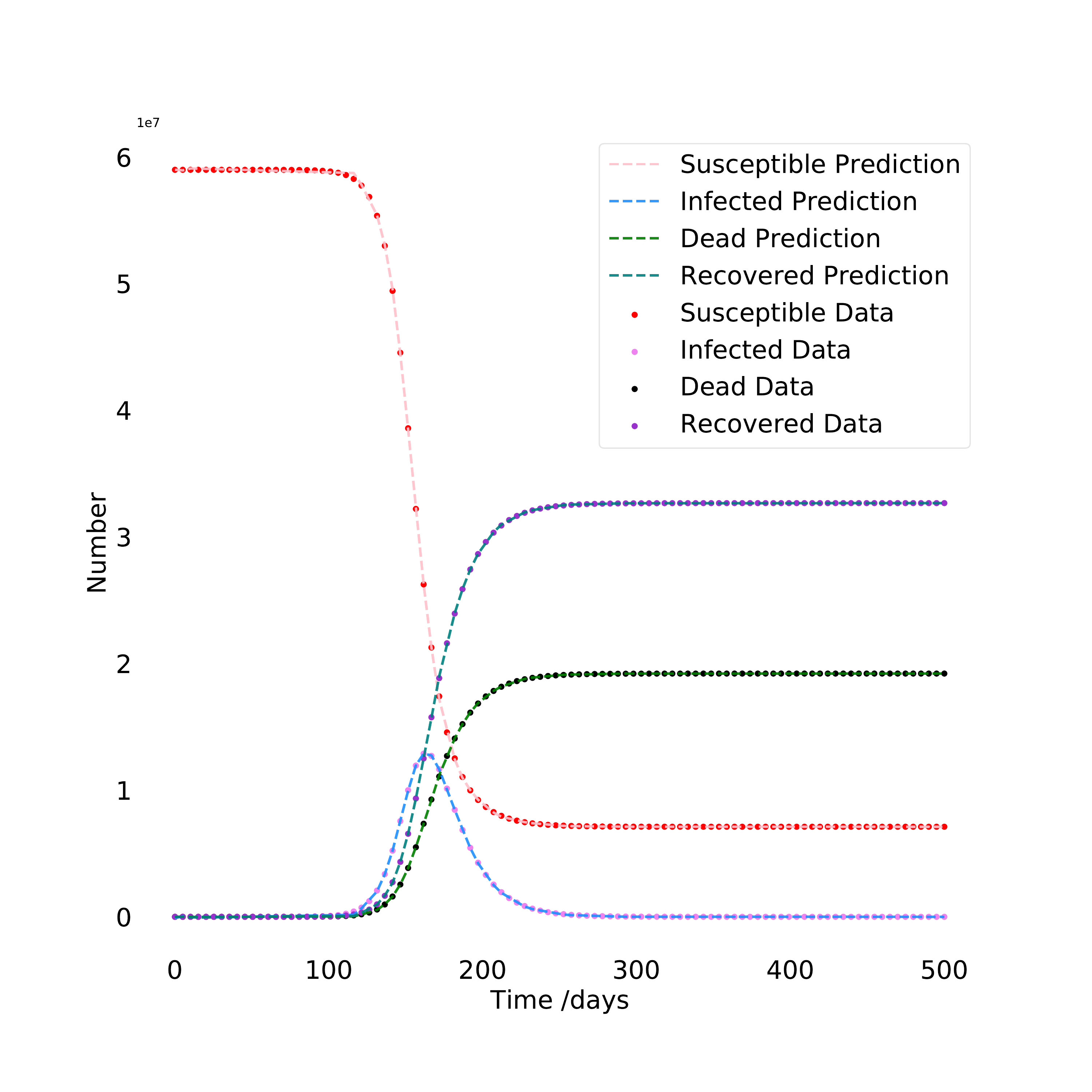}
\caption{100\% search range for NN Output (Left panel) vs LSODA generation from Learnable Parameters (Right Panel)}
\label{100 parameters ranges figure}
\end{figure}

\begin{figure}[!htb]
\centering
\includegraphics[width = 2.5in]{images/param_known_nn.pdf} 
\includegraphics[width = 2.5in]{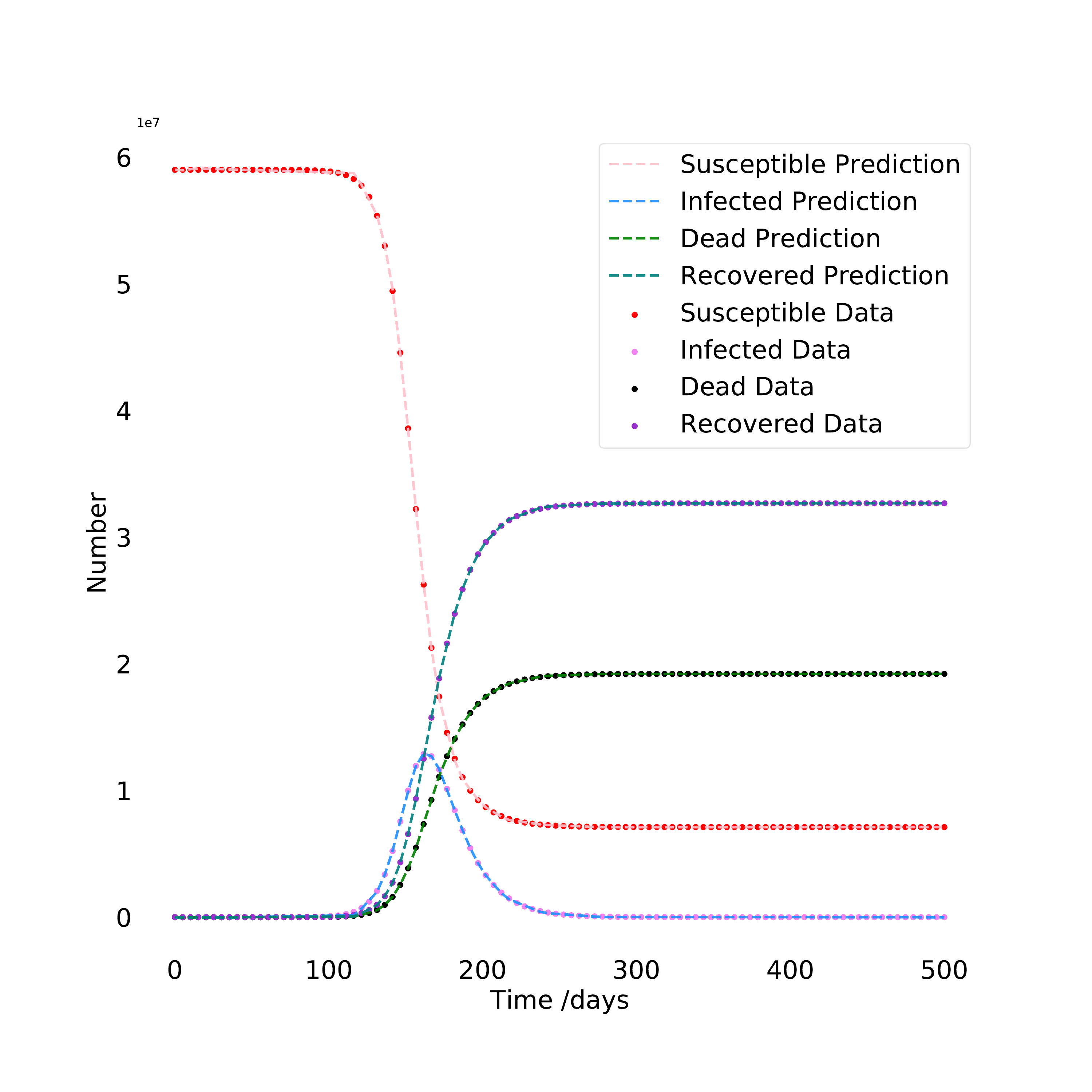}
\caption{1000\% search range for NN Output (Left) vs LSODA generation from Learnable Parameters (Right)}
\label{1000 parameters ranges figure}
\end{figure}

\begin{figure}[!htb]
\centering
\includegraphics[width = 2.5in]{images/param_known_nn.pdf} 
\includegraphics[width = 2.5in]{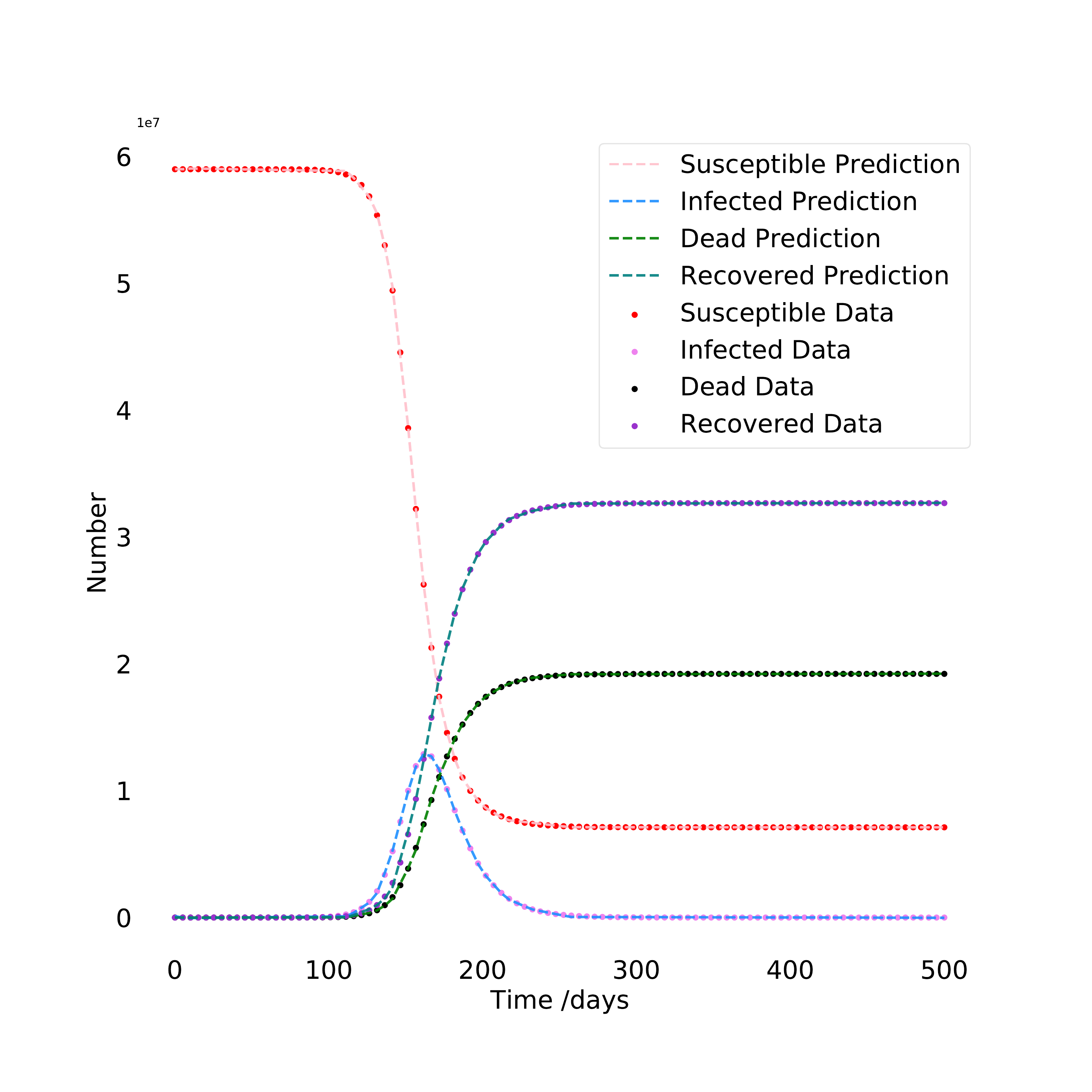}
\caption{10000\% search range for NN Output (Left) vs LSODA generation from Learnable Parameters (Right)}
\label{10000 parameters ranges figure}
\end{figure}

\begin{figure}[!htb]
\centering
\includegraphics[width = 2.5in]{images/param_known_nn.pdf} 
\includegraphics[width = 2.5in]{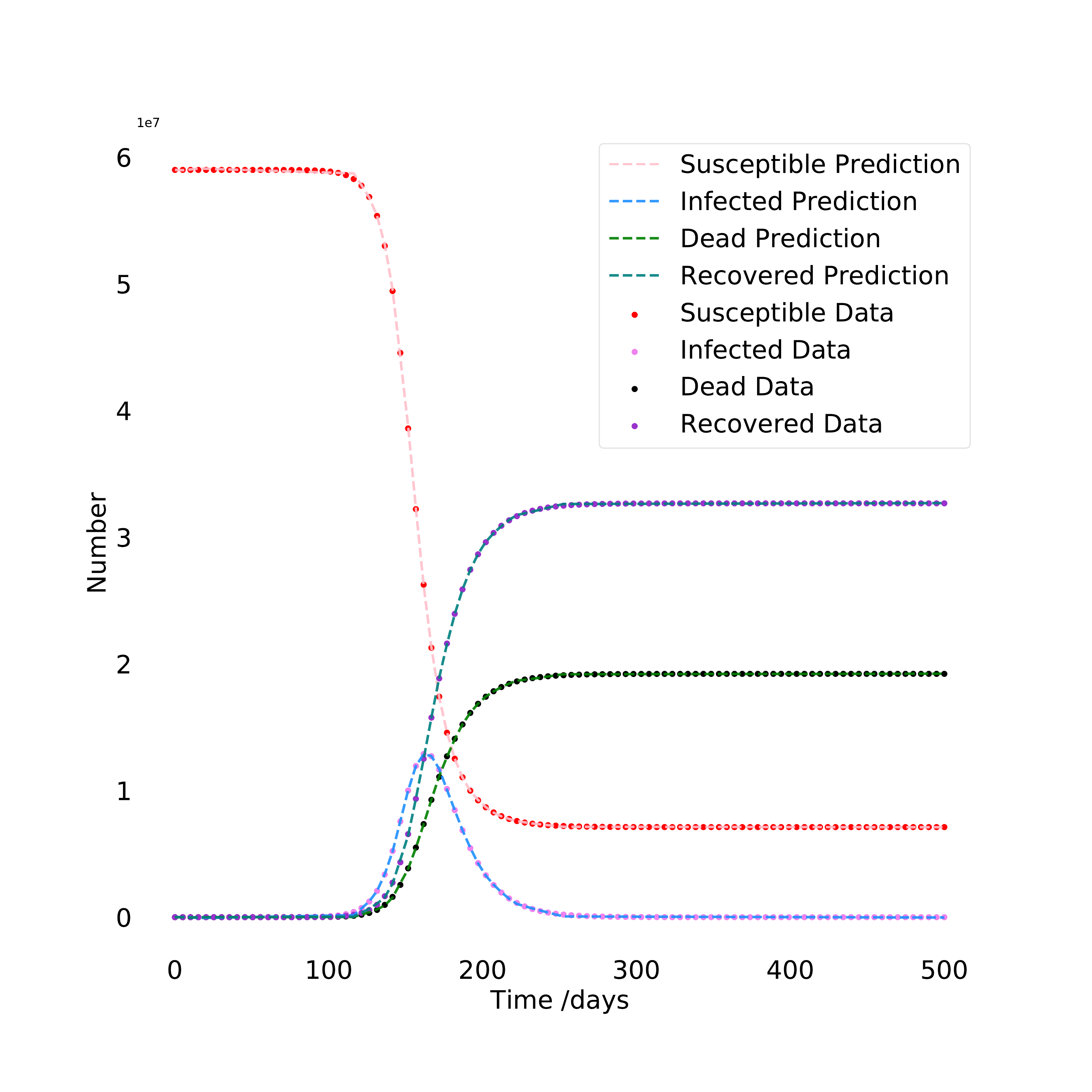}
\caption{100000\% search range for NN Output (Left) vs LSODA generation from Learnable Parameters (Right)}
\label{100000 parameters ranges figure}
\end{figure}

As an example, if the actual parameter's value was $0.1$, a $0\%$ search range would simply be $(0.1, 0.1)$, a $100\%$ range would be $(0.1 + 1 \times 0.1 , -0.1 - 1 \times0.1) = (0.2, - 0.2) $. Further ranges are multiplications of those: $1000\% = (2, -2)$, $10000\% = (20, -20)$, and so on. That is, each unknown parameter is initialized with a random initial value in between its corresponding search range.
\begin{table}[h!]
    \caption{Parameter predictions and relative MSE loss errors for various ranges}
    \centering
    \includegraphics[width=0.9\columnwidth]{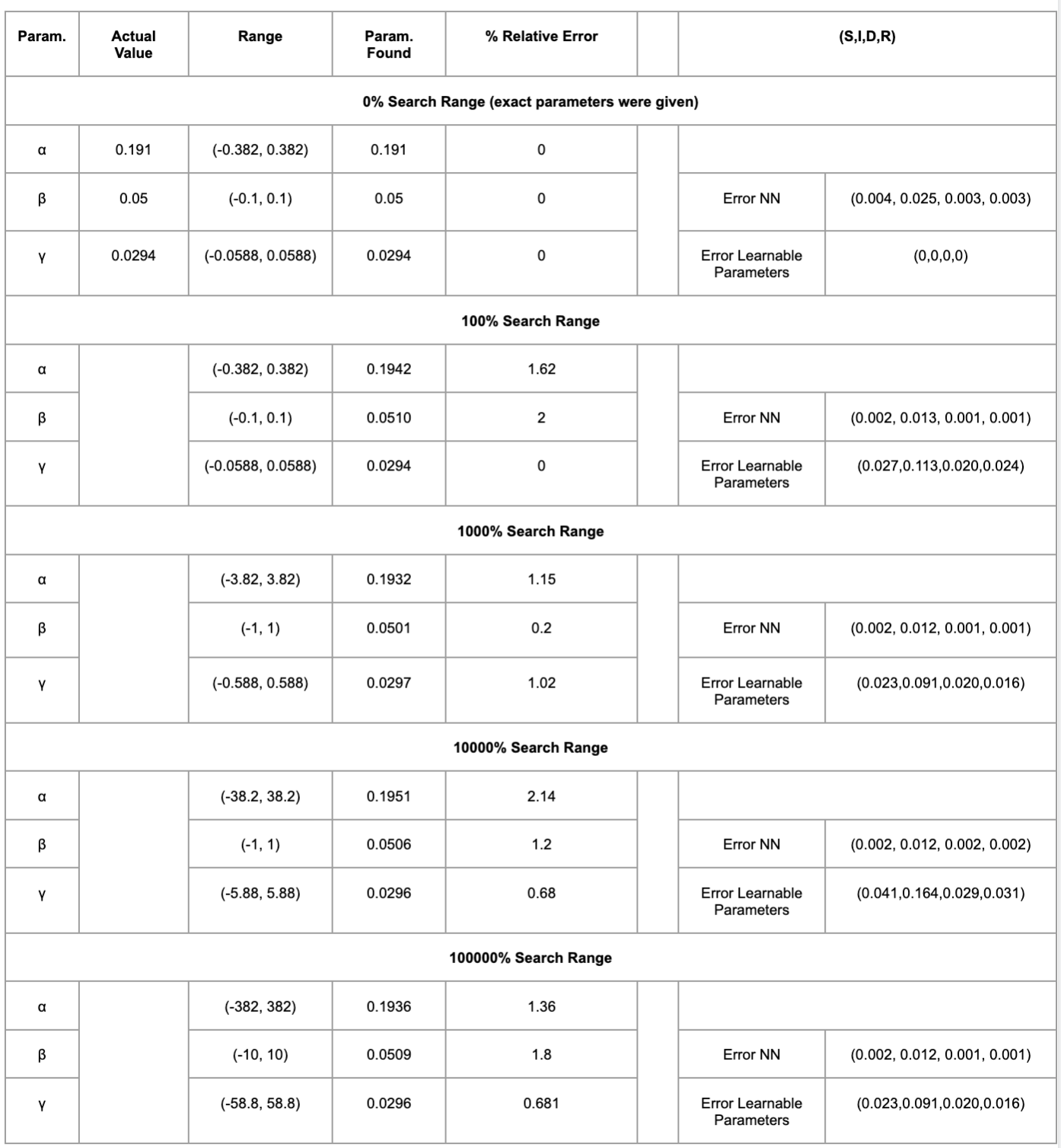}
\label{all_ranges_table}
\end{table}
Table \ref{all_ranges_table} (left) shows the parameters, their actual value, the range DINNs was searching in, and the parameters values that were found by DINNs. The right part of the table shows the error of the neural network and the LSODA generation of the system from the parameters. That is, it shows the effect that the search range had on how well the neural networks' learned the parameters. As seen from table \ref{all_ranges_table} and figures \ref{0 parameters ranges figure} - \ref{100000 parameters ranges figure}, 
at least in the case of the COVID-19 system (\ref{eqnSIRD-S})-(\ref{eqnSIRD-D}), DINNs managed to find extremely close set of parameters in any range we tested. Specifically, in figures \ref{0 parameters ranges figure} - \ref{100000 parameters ranges figure}, the panel on the left shows the effect that the parameter search range had on the neural networks' outputs and the right panel results show the effect that the search ranges had on how well the neural networks' learned the parameters. Additionally, the systems were almost learned perfectly, though, there was some variation in the relative error between experiments. It is worth noting that DINN might be able to learn the system quite well while also having some discrepancies in the learned parameters. Several reasons can explain this such as having a disease system that is relatively simple to learn and a too complex deep learning network, or that DINN found another set of parameters that can explain the data.


\subsubsection{Influence of Noise}
\begin{figure}[h!]
\centering
\subfloat[1\% --- Neural Network's System]{\includegraphics[width = 3.25in]{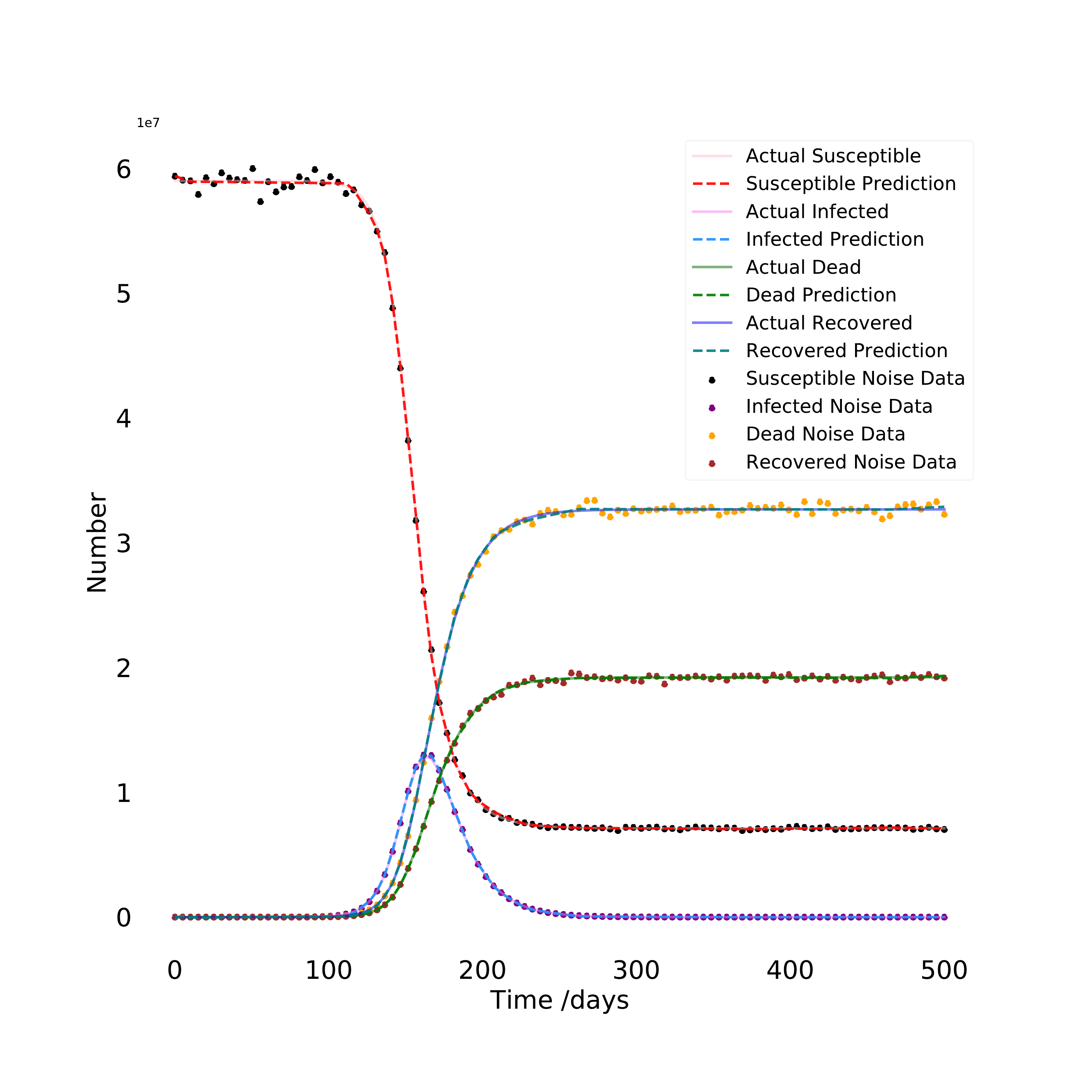}} 
\subfloat[5\% --- Neural Network's System]{\includegraphics[width = 3.25in]{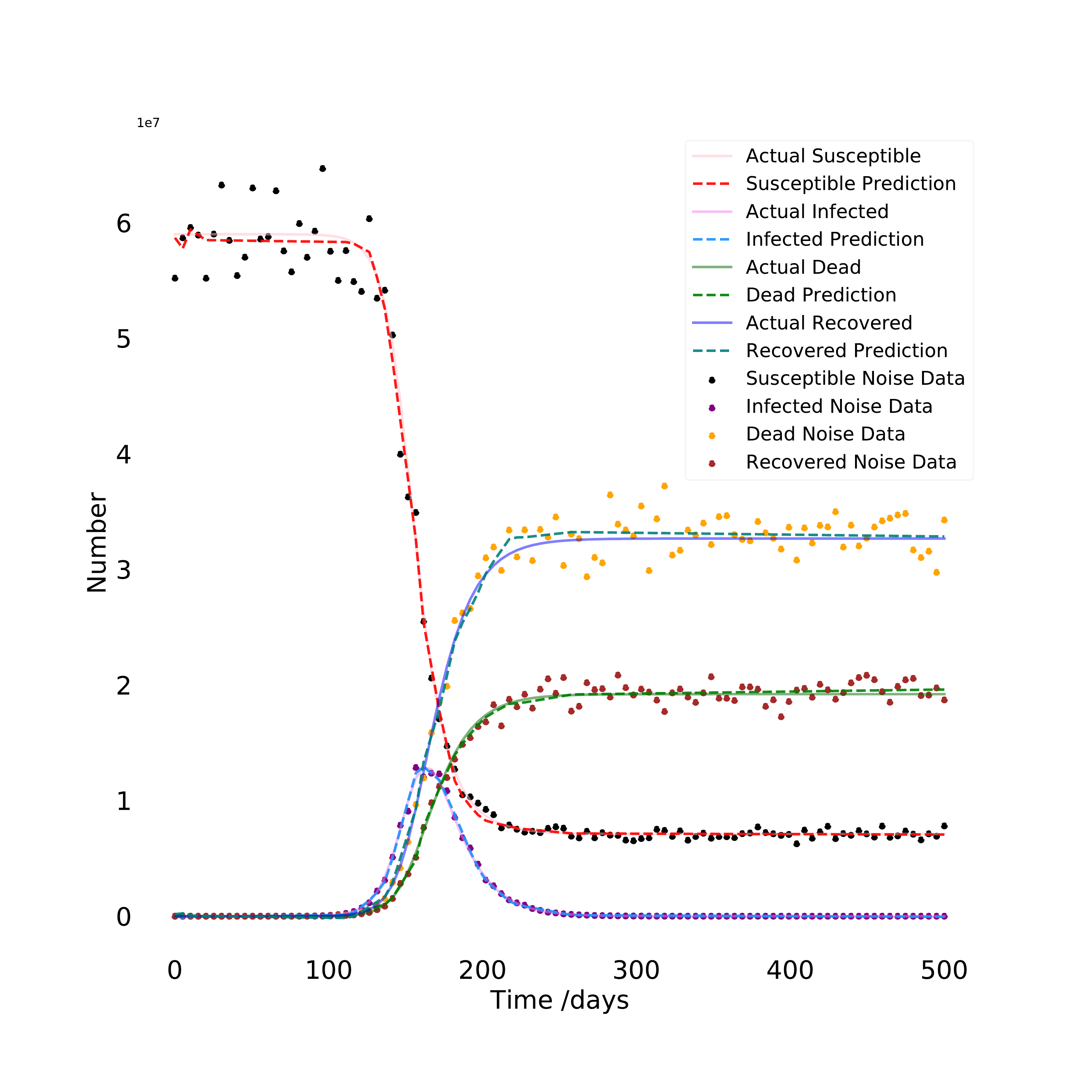}}\\
\subfloat[10\% --- Neural Network's System]{\includegraphics[width = 3.25in]{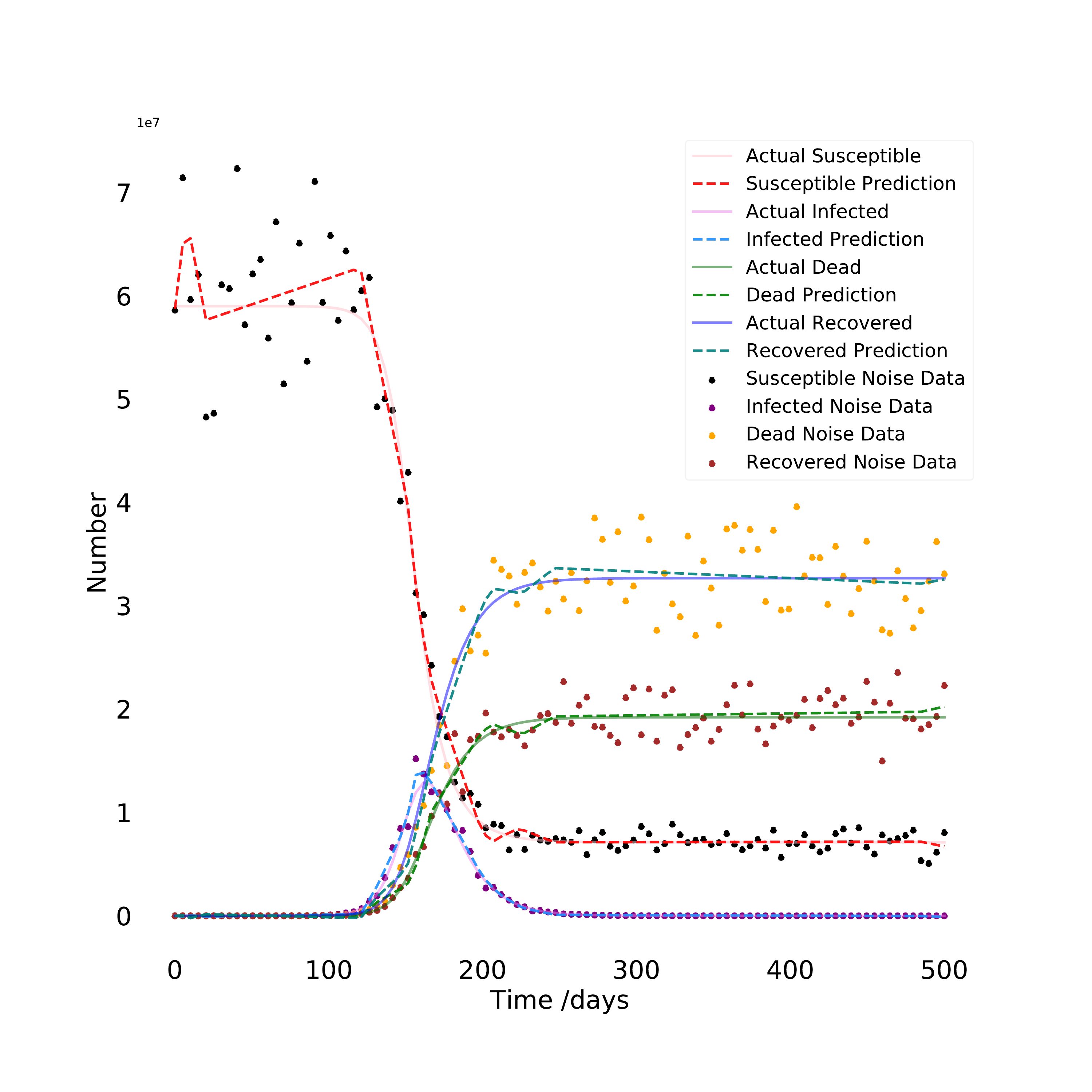}}
\subfloat[20\% --- Neural Network's System]{\includegraphics[width = 3.25in]{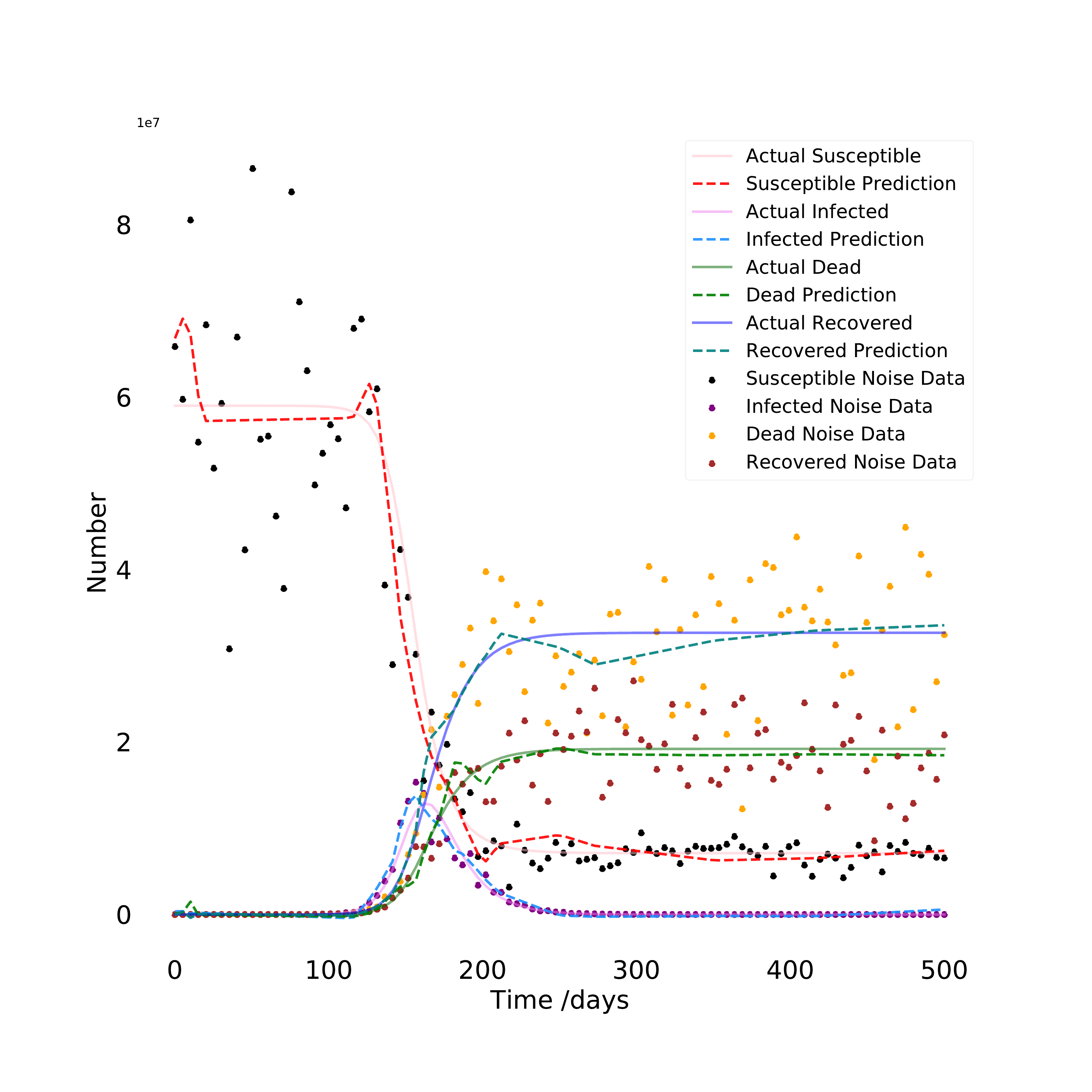}} 
\caption{DINNs performance with varying Uncorrelated Gaussian Noise}
\label{vary percent noise2}
\end{figure}

Next, to show the robustness of DINNs, we generated various amounts of uncorrelated Gaussian noise. The models were trained for $1.4$ million iterations (roughly 1 hour), using parameter ranges of 1000\% variation and similar learning parameters (e.g., learning rate) as the previous section. We used a 4 layer neural network with 20 neurons each, and 100 data points. The experiments showed that even with a very high amount of noise such as 20\%, DINNs achieves accurate results with maximum relative error of $0.143$ on learning the system. That being said, the exact parameters were harder to learn in that amount of noise. It appears that the models may need further training to stabilize the parameters, as there were some variations in the amount of noise versus the accuracy. Figure \ref{vary percent noise2} shows DINN's predictions on 1\%, 5\%, 10\% and 20\% uncorrelated gaussian noise respectively. Table \ref{noises_table} summarizes the estimated optimal parameters for these varying noises.

\begin{table}[t!]
    \caption{Parameter Values for various Uncorrelated Gaussian Noises}
    \centering
    \includegraphics[width=6in]{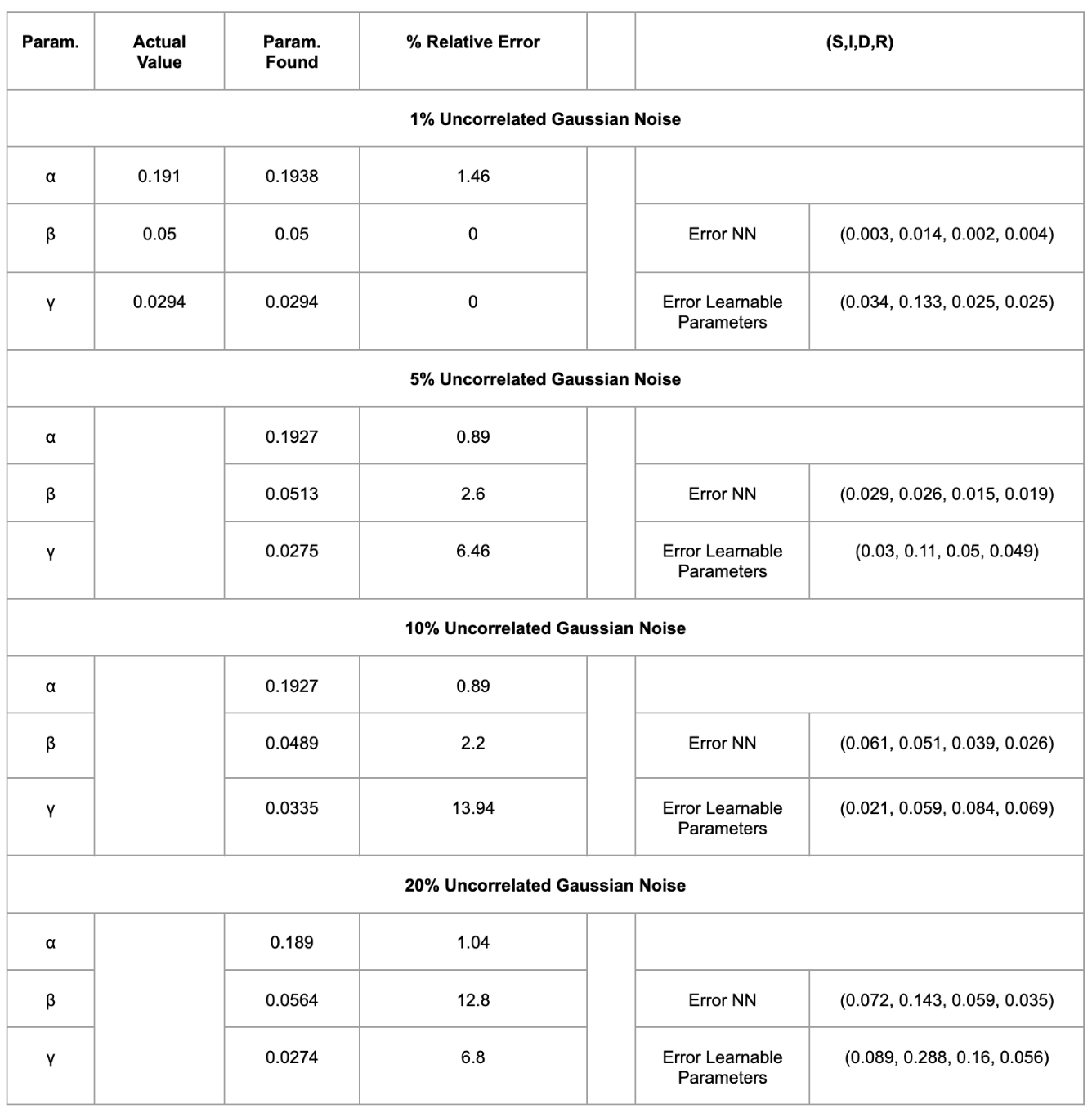}
\label{noises_table}
\end{table}


\subsubsection{Influence of Data Variability}
\begin{figure}[h!]
\centering
\subfloat[10 points --- Neural Network's System]{\includegraphics[width = 3.25in]{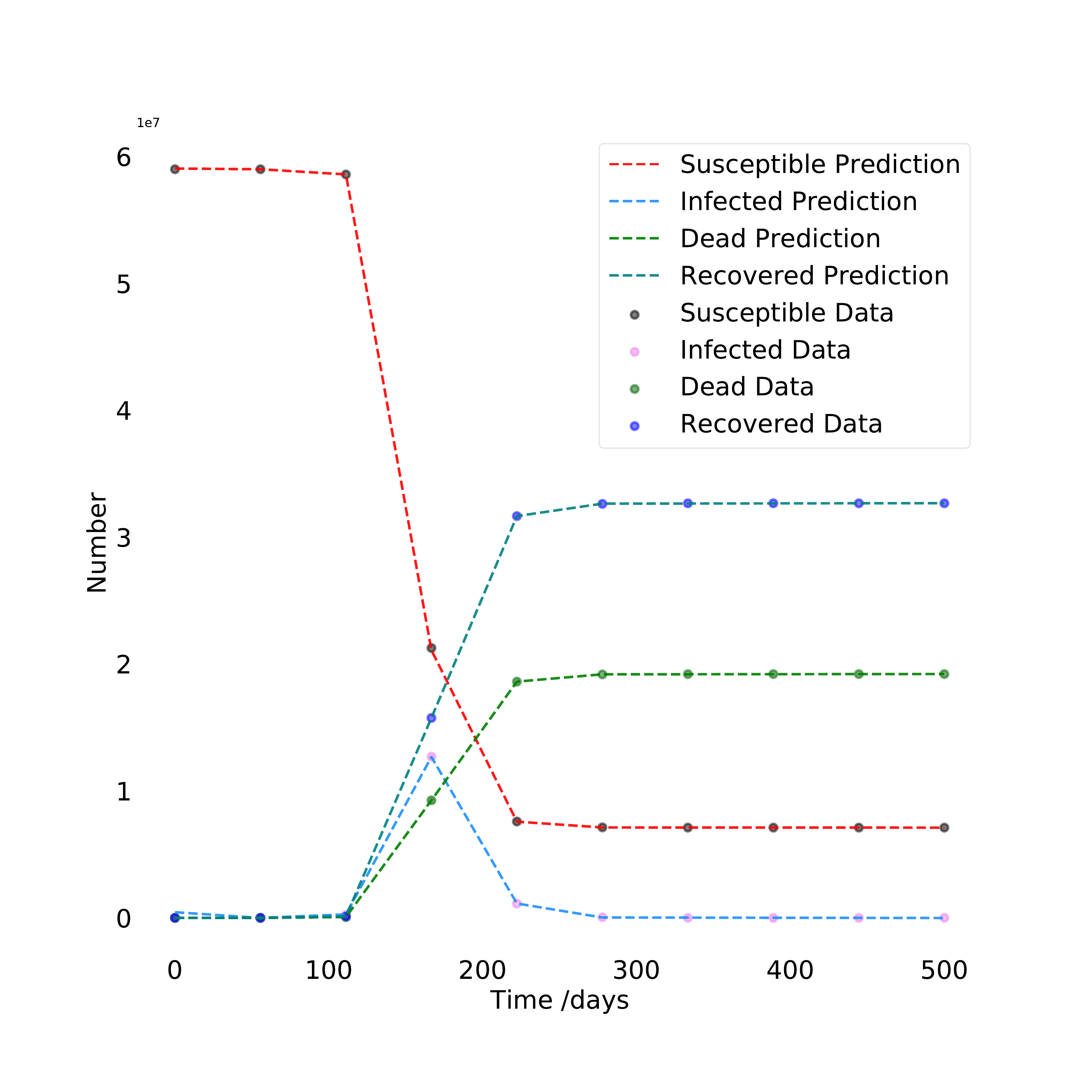}} 
\subfloat[20 points --- Neural Network's System]{\includegraphics[width = 3.25in]{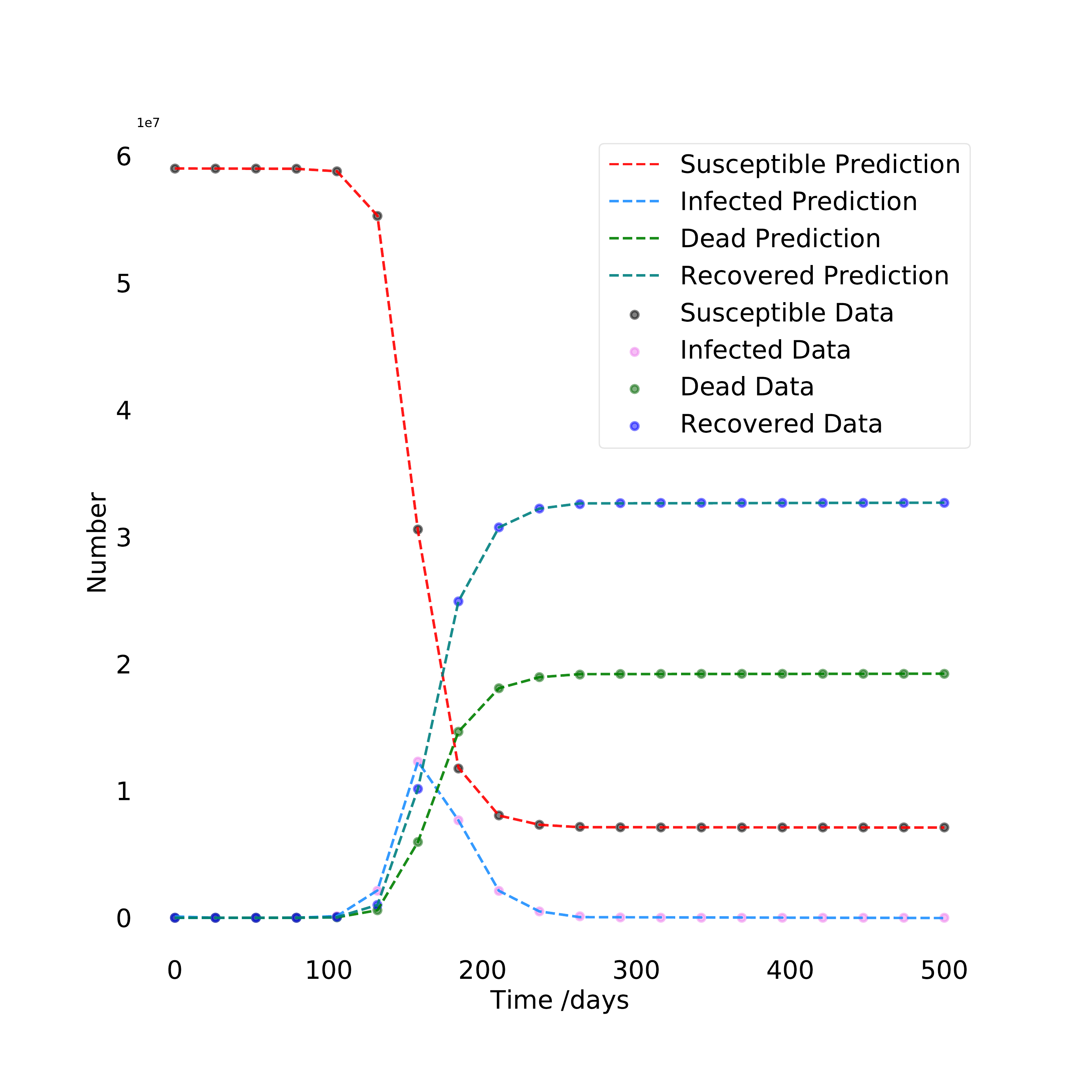}}\\
\subfloat[100 points --- Neural Network's System]{\includegraphics[width = 3.25in]{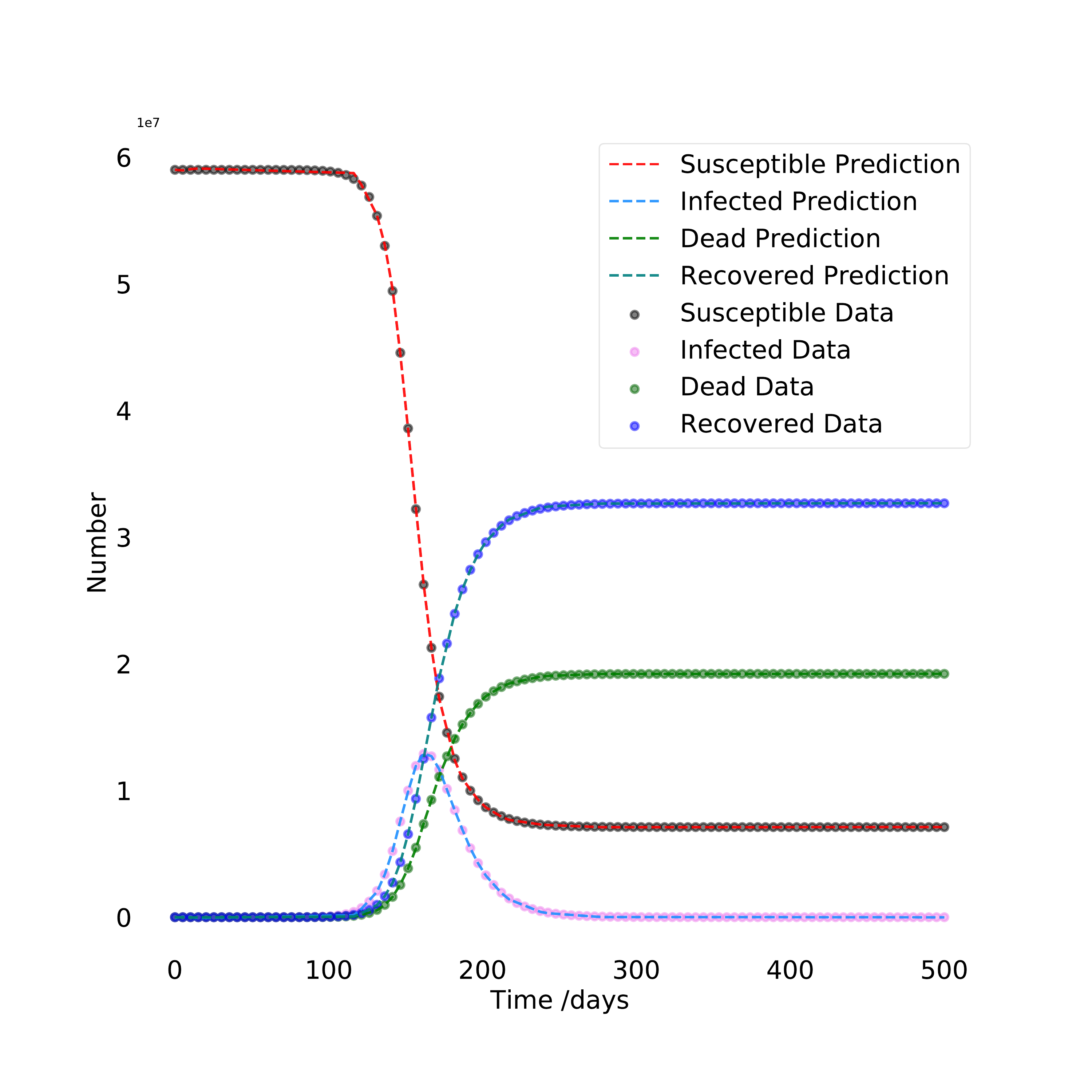}} 
\subfloat[1000 points --- Neural Network's System]{\includegraphics[width = 3.25in]{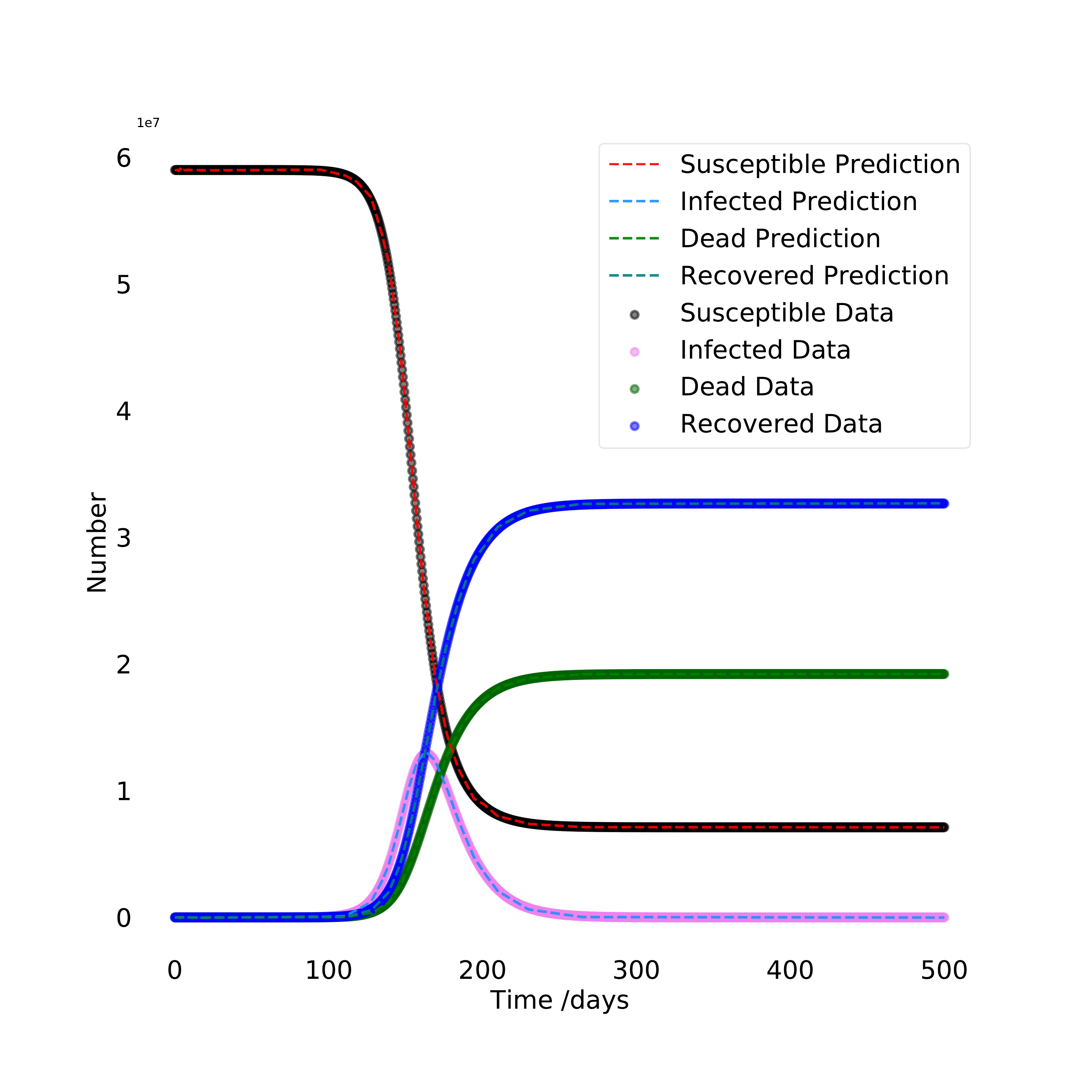}} 
\caption{DINNs performance for increasing Data Points: 10 (top left), 20(top right), 100(bottom left), 1000,(bottom right) }
\label{data_variability_fig}
\end{figure}

\begin{figure}[h!]
\centering
\includegraphics[height=2in,width = 3.25in]{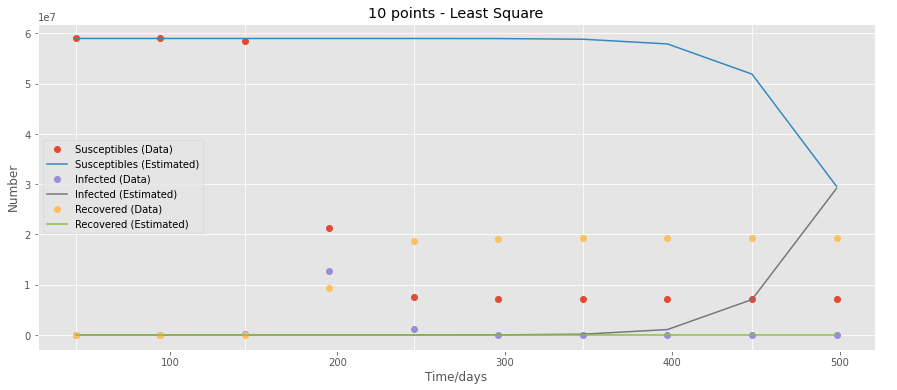}
\includegraphics[height=2in,width = 3.25in]{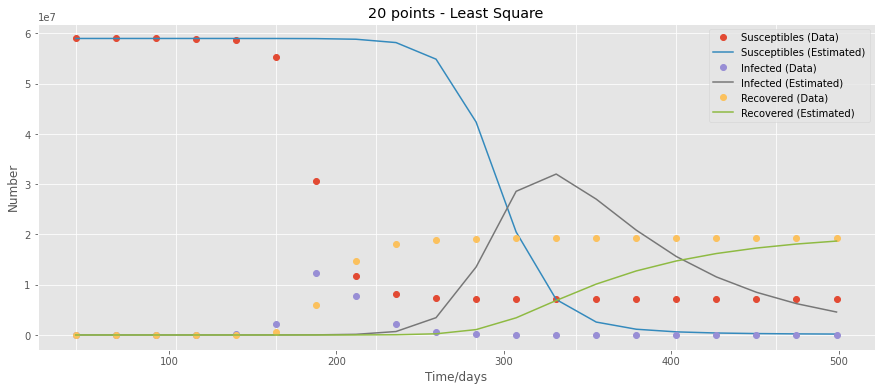}\\
\includegraphics[height=2in,width = 3.25in]{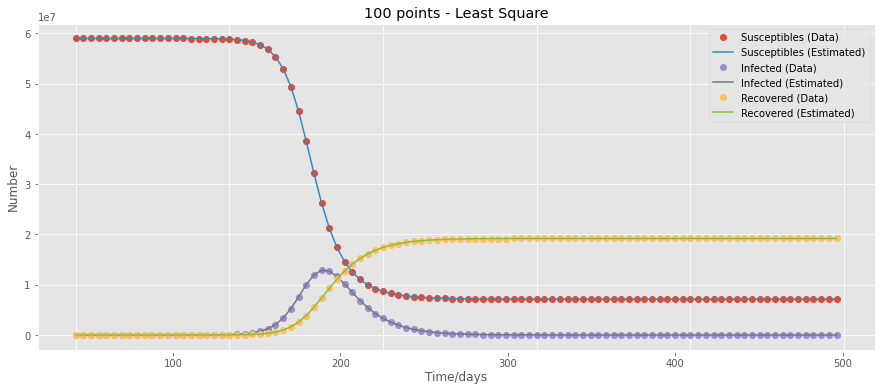} 
\includegraphics[height=2in,width = 3.25in]{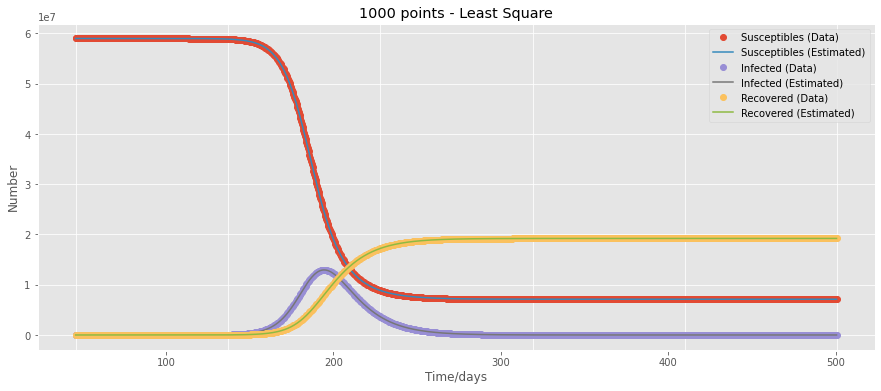}
\caption{Least-squares performance for increasing Points: 10 (top left), 20(top right), 100(bottom left), 1000,(bottom right) }
\label{data_variability_fig2}
\end{figure}

\begin{figure}[h!]
\centering
\includegraphics[height=2in,width = 3.25in]{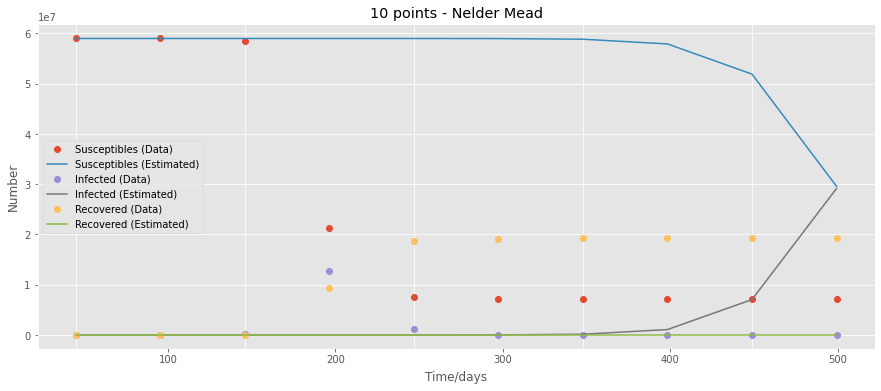} 
\includegraphics[height=2in,width = 3.25in]{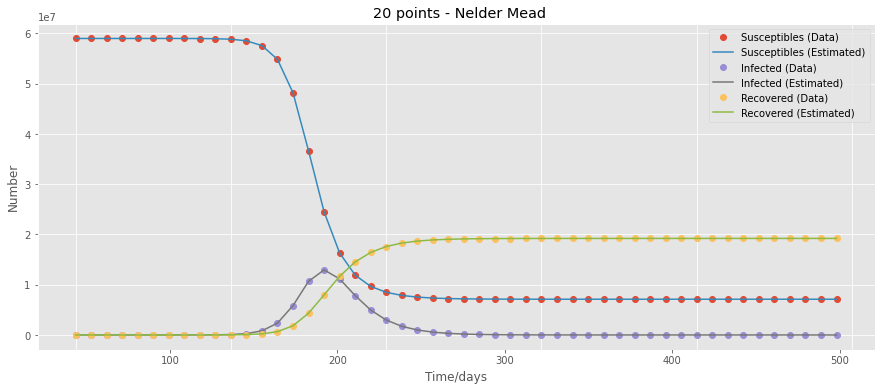}\\
\includegraphics[height=2in,width = 3.25in]{images/100_SIRD_DATA.png} 
\includegraphics[height=2in,width = 3.25in]{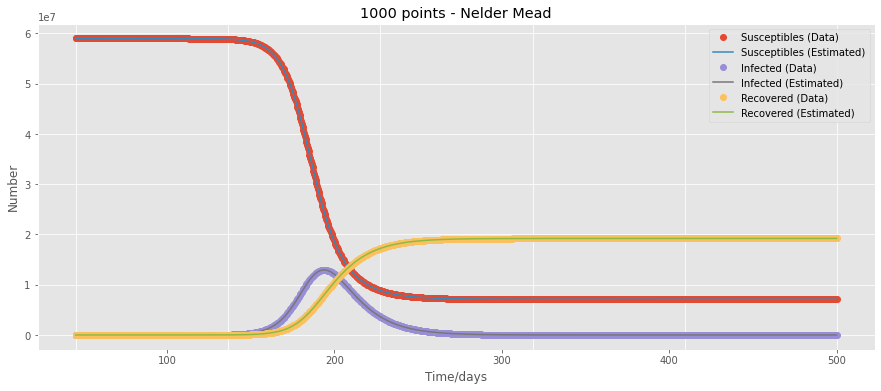} 
\caption{Nelder-Mead performance for increasing Points: 10 (top left), 20(top right), 100(bottom left), 1000,(bottom right) }
\label{data_variability_fig3}
\end{figure}

Next, we trained our models with various amounts of data --- 10, 20, 100, and 1000 points (See Figure \ref{data_variability_fig}). The models were trained for 700,000 iterations, consisting of 4 layers with 20 neurons each, and $1 \times 10^{-6}$ learning rate. Our analysis shows that there was a big increase in the parameters accuracy from 10 points to 20 points. The model that was trained on 1000 data points performed the best compared to the others. 
Note that even with 20 data points the model learns the system incredibly well (See Table \ref{20 data points table}). The left-hand side of the table shows the parameters and values found after training. The right-hand side as before shows the two errors: “Error NN" is the relative MSE loss error from the system that the neural network output (what DINNs believes the systems' dynamics look like), and  “Error Learnable Parameters" is the relative MSE loss error from the LSODA generated system using the parameters found values. DINNs was also compared against a traditional least-squares approach using Gauss-Newton and the Nelder-mead method with variable data points. Both of these techniques require an initial guess for the parameters which was chosen to be $(0.1, 0.1, 0.1)$. Additionally, a search range which the algorithms could search for parameters within was also included to be $(0, 2)$. The results of the two traditional approaches are illustrated in Figure \ref{data_variability_fig2} and Figure \ref{data_variability_fig3} respectively, which show that these traditional approaches only start to perform comparable to DINNs when there are more data points. While Nelder-Mead was a little better than Gauss-Newton method, both could not perform as well as DINNs for a minimal dataset. More experiments were also conducted by increasing the search range for the parameters, but both the Gauss-Newton and Nelder-Mead performed worse. 
\begin{table}[h!]
\caption{Performance of DINNs for 20 data points}
    \centering
    \includegraphics[width=6in]{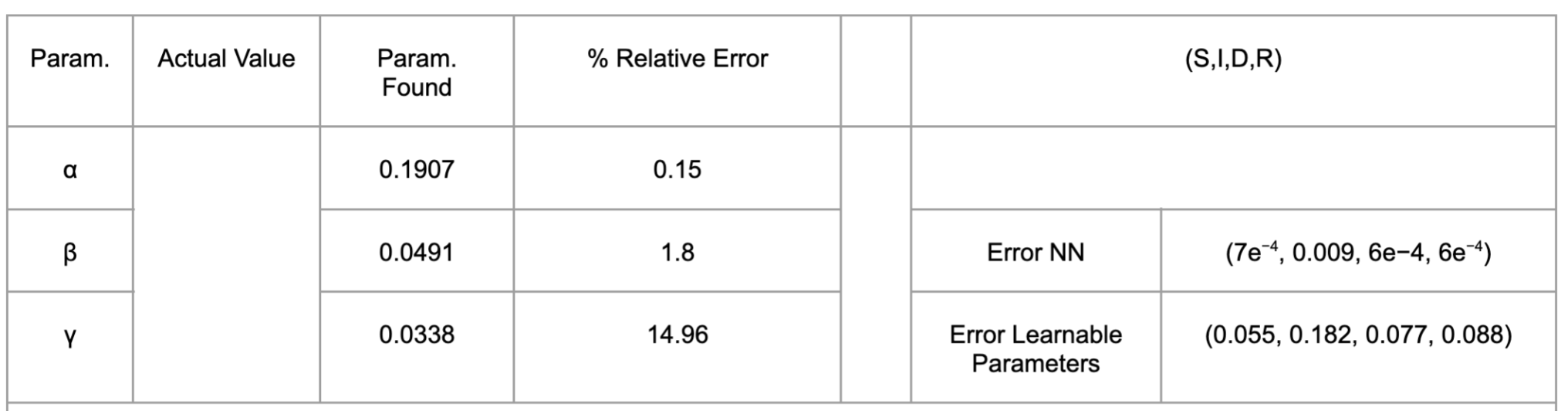}
\label{20 data points table}
\end{table}

\subsubsection{Influence of Neural Network Architectures}
In the next computational experiment, we examined the effect that wider or deeper neural network architecture has on DINNs. The models were trained on 100 data points, using parameter ranges of 1000\%, a learning rate of $1 \times 10^{-6}$, and 700,000 iterations. Tables \ref{NN Architecture} and \ref{Parameters Architecture} show a clear decrease in error as one increases the amount of neurons per layer. Specifically, Table \ref{NN Architecture} itemizes the (S,I,D,R) error from the neural network's output. For the Neural network architecture variations (depth and width), relative MSE errors were reported on the predicted NN system. Table \ref{Parameters Architecture} itemizes similar findings for LSODA generation of the learning parameters. There also seem to be a clear decrease in error as the number of layers increase. However, the error seem to stabilize around 8 layers, with very minor performance increase in 12 layers.
\begin{table}[h!]
\caption{Influence of Architecture Variation with (S, I, D, R) error from neural network output}
\label{tab:neurons}
\centering
\renewcommand\arraystretch{1.1}
\begin{tabular}{@{} *{4}{c} @{}}
    \toprule
    &   \multicolumn{3}{c}{Neurons Per Layer} \\
    \cmidrule(l){2-4}
Layers
    & 10 & 20 & 64 \\
    \midrule
    2 & (0.030, 0.109, 0.027, 0.027) & (0.024, 0.15, 0.022, 0.022) & (0.005, 0.038, 0.004, 0.004) \\
    4 & (0.002, 0.027, 0.004, 0.004) & (0.001, 0.007, 0.001, $8e^{-4}$) & ($8e^{-4}$, 0.004, $7e^{-4}$, $7e^{-4}$) \\
    8 & (0.001, 0.008, 0.002, 0.002) & ($4e^{-4}$, 0.002, $5e^{-4}$, $4e^{-4}$) & ($3e^{-4}$, 0.001, $2e^{-4}$, $1e^{-4}$) \\
    12 & (0.001, 0.008, 0.002, 0.002) & ($5e^{-4}$, 0.002, $8e^{-4}$, $6e^{-4}$) & ($2e^{-4}$, 0.001, $2e^{-4}$, $2e^{-4}$) \\
        \bottomrule
\end{tabular}
\label{NN Architecture}
    \end{table}

\begin{table}[h!]
\caption{Influence of Architecture Variation with (S, I, D, R) error from LSODA}
\label{tab:neurons}
\centering
\renewcommand\arraystretch{1.1}
\begin{tabular}{@{} *{4}{c} @{}}
    \toprule
    &   \multicolumn{3}{c}{Neurons Per Layer} \\
    \cmidrule(l){2-4}
Layers
    & 10 & 20 & 64 \\
    \midrule
    2 & (0.132, 0.519, 0.088, 0.111) & (0.106, 0.423, 0.083, 0.077) & (0.001, 0.009, 0.019, 0.011) \\
    4 & (0.038, 0.148, 0.026, 0.029) & (0.064, 0.256, 0.045, 0.050) & (0.009, 0.044, 0.010, 0.008) \\
    8 & (0.036, 0.138, 0.033, 0.024) & (0.027, 0.107, 0.018, 0.022) & (0.057, 0.234, 0.045, 0.043) \\
    12 & (0.036, 0.138, 0.033, 0.024) & (0.022, 0.091, 0.015, 0.019) & (0.017, 0.076, 0.017, 0.017) \\
        \bottomrule
\end{tabular}
\label{Parameters Architecture}
    \end{table}
    
\subsubsection{Influence of Learning Rates}
We found that quickly increasing the learning rates and then quickly decreasing it to a steady value allows the network to learn well. One such learning rate schedule is PyTorch’s CyclicLR learning rate scheduler.
To show the importance of learning rate in the amount of needed training time, we trained DINNs with several values: $1\times 10^{-5}$, $1\times 10^{-6}$, $1\times 10^{-8}$ as well as different step size for each one: 100, 1000, 10000. We used 4 layers with 20 neurons each, and 100 data points. The time measured from the moment the network started training, and until the loss was smaller than $4 \times 10^{-4}$ --- which usually corresponds to learning the system almost perfectly. As can be seen from the results (Table \ref{learning rate table}) both the minimum learning rate and the step size play an important role in learning the system. Reducing the learning rate to a small value too quickly may result in hours of training time instead of minutes. As an afterthought, this might be the reason why most of the systems were taking so long to train (\textgreater 10 hrs), while the COVID system took \textless 25 minutes.
\begin{table}[h!]
\caption{Learning Rate \& Step Size vs Training Time}
\label{tab:neurons}
\centering
\renewcommand\arraystretch{1.1}
\begin{tabular}{@{} *{4}{c} @{}}
    \toprule
    &   \multicolumn{3}{c}{Step Size Up} \\
    \cmidrule(l){2-4}
Learning Rate
    & 100 & 1000 & 10000 \\
    \midrule
    $1\times 10^{-5}$ & 2min 31s & 2min 57s & 3min 16s \\
    $1\times 10^{-6}$ & 21min 11s & 20min 59s & 18min 43s \\
    $1\times 10^{-8}$& \textgreater8hrs & \textgreater8hrs & \textgreater8hrs \\
        \bottomrule
\end{tabular}
\label{learning rate table}
    \end{table}
    
\subsubsection{Application of DINNs to Real-Data}
Finally, to verify that DINNs is in fact as reliable as it appears, we used 310 days (04-12-2020 to 02-16-2021) of real US data from \cite{JHU}. We trained a neural network that learned the cumulative cases of susceptible, infected, dead, and recovered, and predicted the cases for a future month. Specifically, out of those 310 days we gave the network 280 days worth of data and asked it to predict each compartment's progression a month (30 days) into the future. The network received 31 data points (1 per 10 days), was trained for 100k epochs (roughly 5 minutes), had 8 layers with 20 neurons each, a 1000\% parameters variation, and $1 \times 10^{-5}$ learning rate. 

\begin{figure}[t]
    \centering
    \includegraphics[width=0.75\columnwidth]{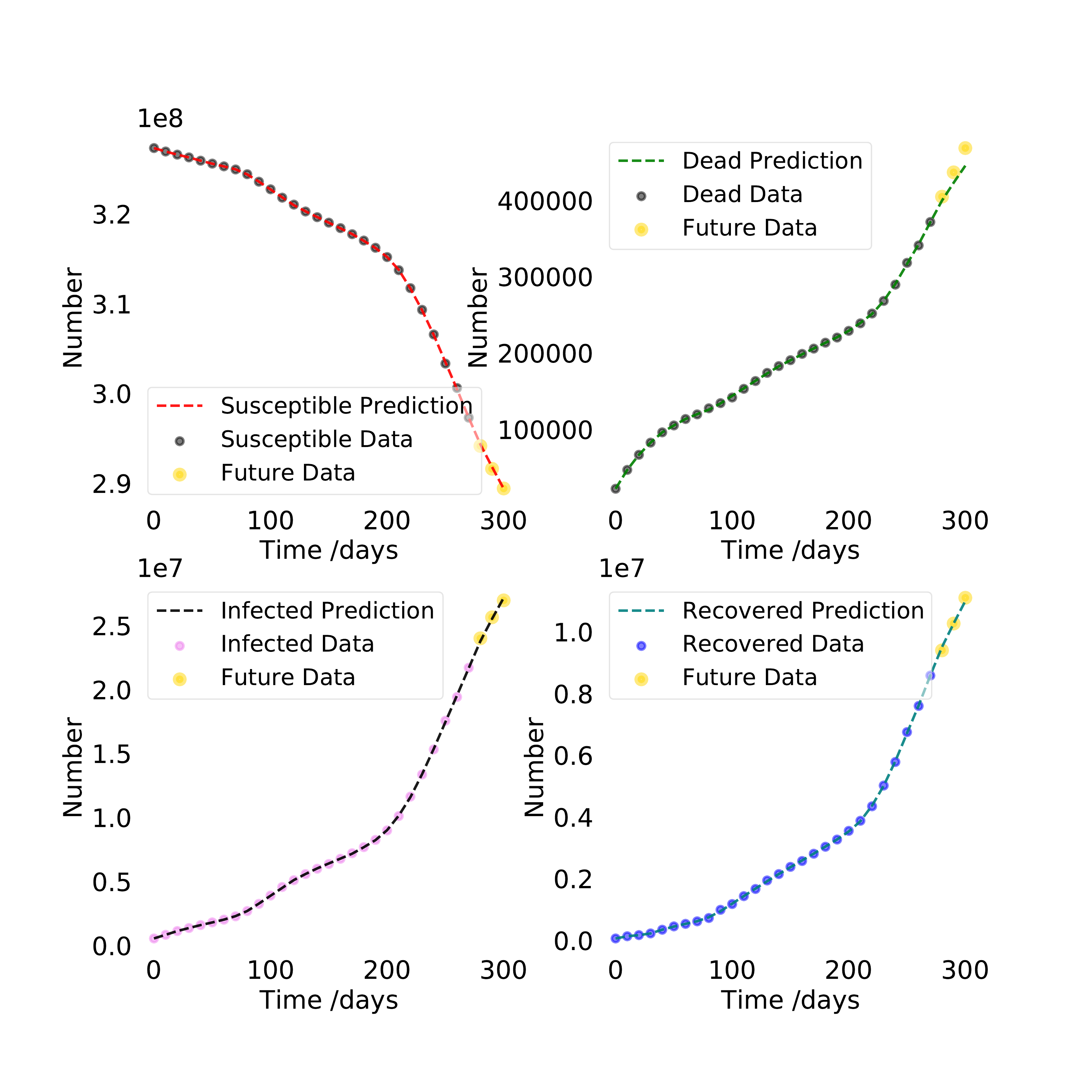}
    \caption{DINN's output on COVID real-life cumulative cases over 310 days}
\label{real_data}
\end{figure}

Our results suggest that the learnable parameters found were quite different from the parameters in the literature ($\alpha=0.0176$ instead of $0.191$, $\beta=0.0046$ instead of $0.05$, and $\gamma=0.0001$ instead of $0.0294$). This may imply that either the data was different from the initial data distribution used in the literature \cite{Anastassopoulou2019}, or that as other authors mentioned these are time-varying parameters rather than constant ones. As seen from figure \ref{real_data}, the cumulative cases had less data variation and were fairly easy to learn. Additionally, it appears as DINNs managed to accurately predict the future month on each compartment. 

\subsubsection{Influence of Missing Data}
\begin{figure}[h!]
\centering
\includegraphics[width = 2.8in]{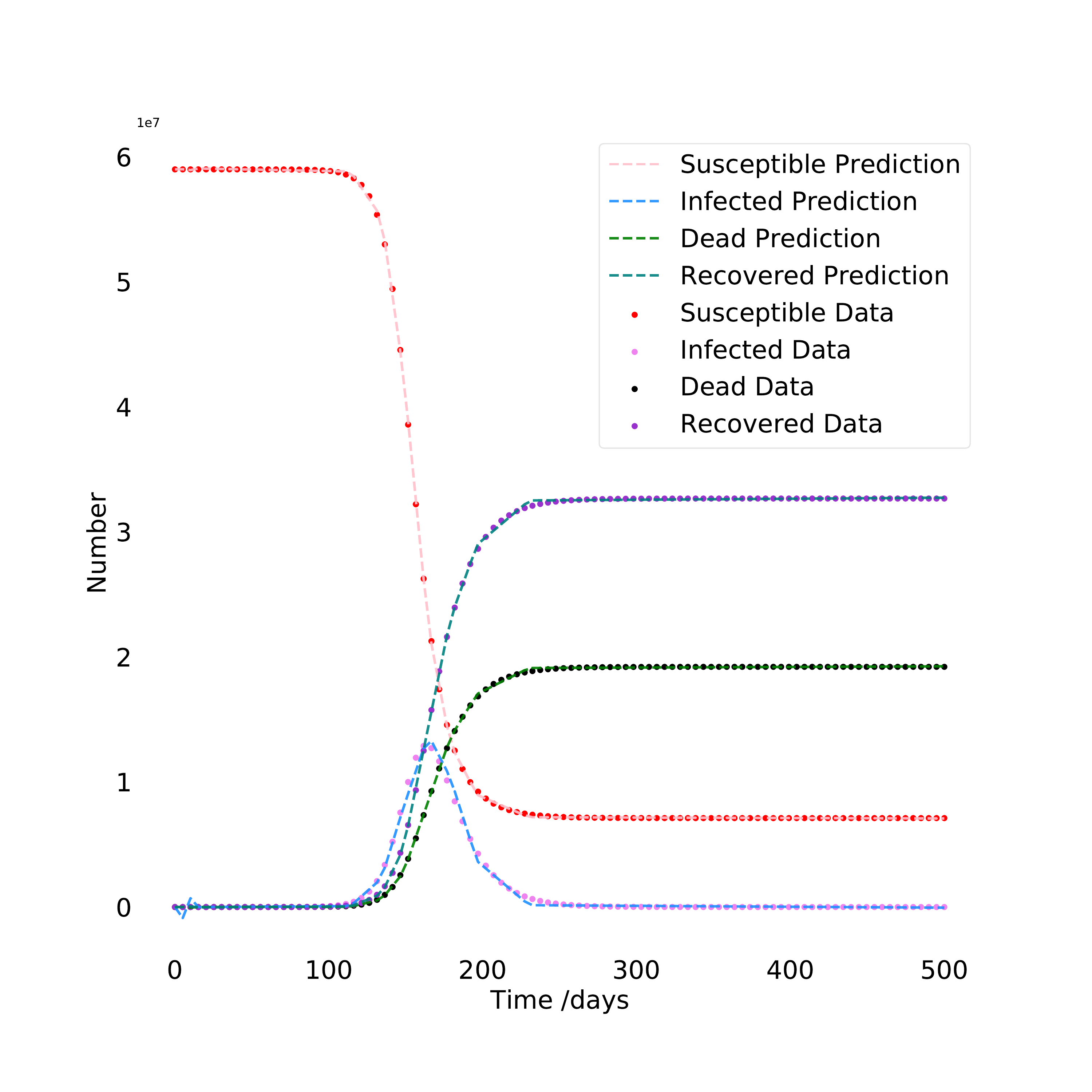}
\caption{Performance of DINNs on Missing data for COVID}
\label{fig: missing data2}
\end{figure}

So far we assumed that we have all the data for each compartment. However, this is often not the case. For example, there is a lot of data that went unreported during COVID-19. To test the reliability of DINNs, we tested the method on the SIRD model again which was trained on 100 data points, were given the known parameters from the literature, and were only given the initial conditions for the missing data. The model was trained with $1\times 10^{-6}$ learning rate for 1 million iterations (roughly 1 hour). 
Our results show that DINNs can in fact learn systems even when given partial data. However, it is important to note that the missing data compartments should be in at least one other compartment in order to get good results. For example, when we tried to remove the COVID recovered compartment (i.e., R), DINNs learned S, I, and D nearly perfectly. However, it did not do very well on R. That is because R is not in any of the other equations. The neural networks' systems outputs and their losses for COVID model was $(0.003, \boldsymbol{0.078}, 0.003, 0.003)$. The prediction using these values is shown in Figure \ref{fig: missing data2}.
\begin{figure}[h!]
\centering
\includegraphics[width = 2.8in]{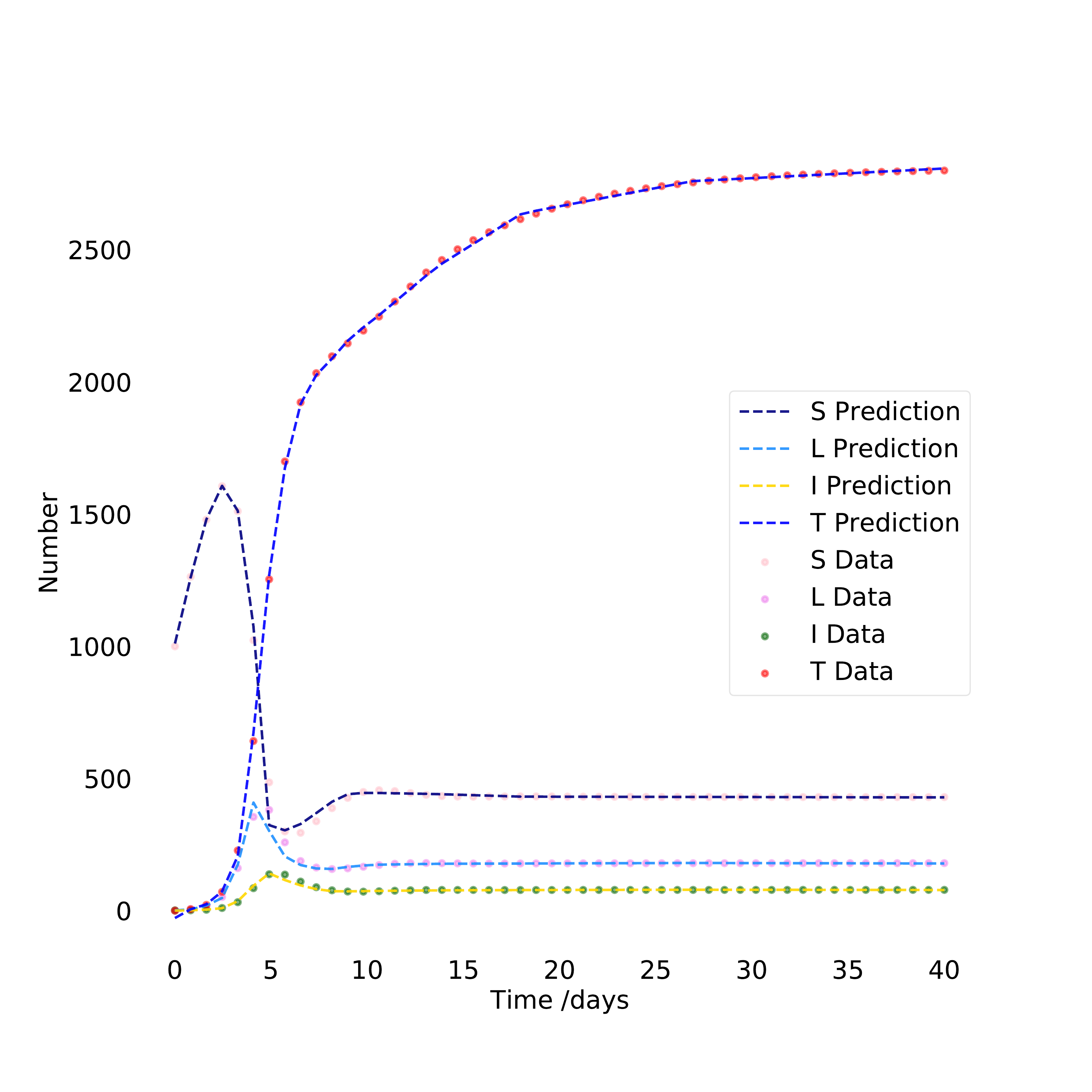}
\includegraphics[width = 2.8in]{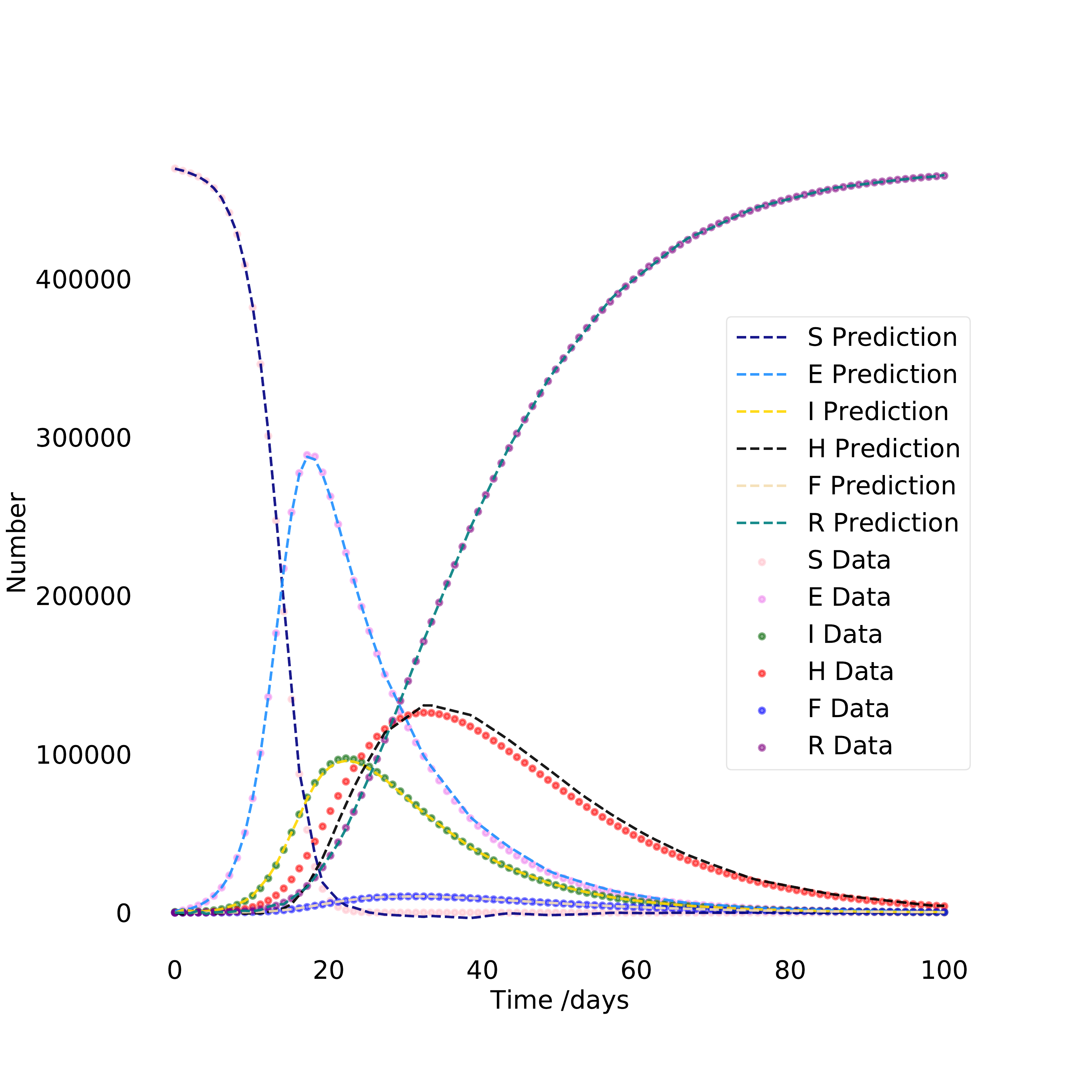}
\caption{Performance of DINNs on Missing data for COVID Tuberculosis (left) and Ebola (right)}
\label{fig: missing data1}
\end{figure}
\subsection{Application of DINNs to other infectious diseases}
In this sections, we apply DINNs to multiple infectious diseases. Note that we chose smaller ranges for the following diseases for demonstrating that DINN can in fact identify the systems and one set of parameters that match the literature they came from, as in many of these systems there exist a large set of parameters that can generate them. However, one can easily expand the ranges as done in a previous section. Similarly to the previous sections we report relative errors, except when the true value of the parameter is zero, which then we use the absolute error.

First, we employ DINNs to two systems modeling diseases with missing data as in the last section. These include,
\begin{itemize}
\item A Tuberculosis SLIT model \cite{castillo1997treat} with missing data on Latent  compartment $L$ and infected individuals $I$ given by:  
\begin{eqnarray*}
dS/dt &=& \delta - \beta  c  S  I / N - \mu  S\\
dL/dt &=& \beta  c  S  I / N - (\mu + k + r_1)  L +
\beta'  c  T /N\\
dI/dt &=& kL - (\mu + d)  I - r_2  I\\
dT/dt &=& r_1  L + r_2  I - \beta'  c  T /N - \mu T
\end{eqnarray*}
\item An Ebola SEIHFR model \cite{legrand2007understanding} with missing data on hospitalized cases $H$ given by:
\begin{eqnarray*}
dS/dt &=& -1/N  (\beta_1  S  I + \beta_h  S  H + \beta_f  S  F)\\
dE/dt &=& 1/N  (\beta_1  S  I + \beta_h  S  H + \beta_f  S  F) - \alpha  E \\
dI/dt &=& \alpha  E - (\gamma_h  \theta_1 + \gamma_i  (1-\theta_1)(1-\delta_1) + \gamma_d  (1-\theta_1)  \delta_1)  I\\
dH/dt &=& \gamma_h  \theta_1  I - (\gamma_dh  \delta_2 + \gamma_ih  (1-\delta_2))  H\\
dF/dt &=& \gamma_d  (1-\theta_1)  \delta_1  I + \gamma_dh  \delta_2  H - \gamma_f  F\\
dR/dt &=& \gamma_i  (1-\theta_1)  (1-\delta_1)  I + \gamma_ih  (1-\delta_2)  H + \gamma_f  F
\end{eqnarray*}
\end{itemize}
The models were also trained on 100 data points, were given the known parameters from the literature, and were only given the initial conditions for the missing data. 
The tuberculosis model was trained with $1\times 10^{-5}$ learning rate for 100k iterations. 
The Ebola model was trained with $1\times 10^{-6}$ learning rate for 800,000 iterations. 
 The neural networks' systems outputs and their losses for the Tuberculosis model was $(0.041, \boldsymbol{0.086}, \boldsymbol{0.051}, 0.004)$  and the Ebola model was $(0.013, 0.011, 0.014, \boldsymbol{0.103}, 0.007, 0.001)$. Figure \ref{fig: missing data2} illustrate the outputs for the respective models.
\subsubsection{A summary of DINNs applied to eleven Diseases}
Expanding on the relatively simple SIRD model for COVID  that was used for simplicity to demonstrate the capability of DINNs, here we apply the method to ten other highly infectious diseases, namely Anthrax, HIV, Zika, Smallpox, Tuberculosis, Pneumonia, Ebola, Dengue, Polio, and Measles. These diseases vary in their complexity, ranging from a system of three to nine ordinary differential equations, and from a few parameters to over a dozen. Table \ref{11_diseases_analysis} provides a summary of our analysis. Specifically, it itemizes for each disease its best, worst, and median parameter estimate error. 
\begin{table}[H]
\centering
\caption{Summary of the analysis for eleven diseases}
\begin{tabular}[t]{lcccc}
\hline
 Disease & Best & Worse & Median \\
\hline
COVID & 0.2 & 1.151 & 1.02 \\
Anthrax &0.5754 &6.0459 & 2.4492\\
HIV &0.007515 &3.811689 &  0.829756\\
Zika &0.0588 &5.8748 &0.7261 \\
Smallpox &0.0882 &  10.8598& 4.9239\\
Tuberculosis &0.5424 & 11.0583& 3.8952\\
Pneumonia & 0.0005 & 11.6847& 1.6372 \\
Ebola & 0.2565 & 9.6403&1.1244 \\
Dengue &0.2696 &  9.7723& 0.8796\\
Polio &0 & 0.4168& 0.3587\\
Measles &2.9999 &  12.704& 3.1453\\
\hline
\end{tabular}
\label{11_diseases_analysis}
\end{table}
In the subsequent subsections, for each of the diseases described by a system of differential equations, we identify the relative error for the disease from LSODA generation of the learnable parameters, a table representing parameter values (actual and computed values) with their range and percentage relative error, and a graph of the prediction from the data.
\subsubsection{COVID-19}
\begin{figure}[t]%
    \centering
    \includegraphics[width=0.6\columnwidth]{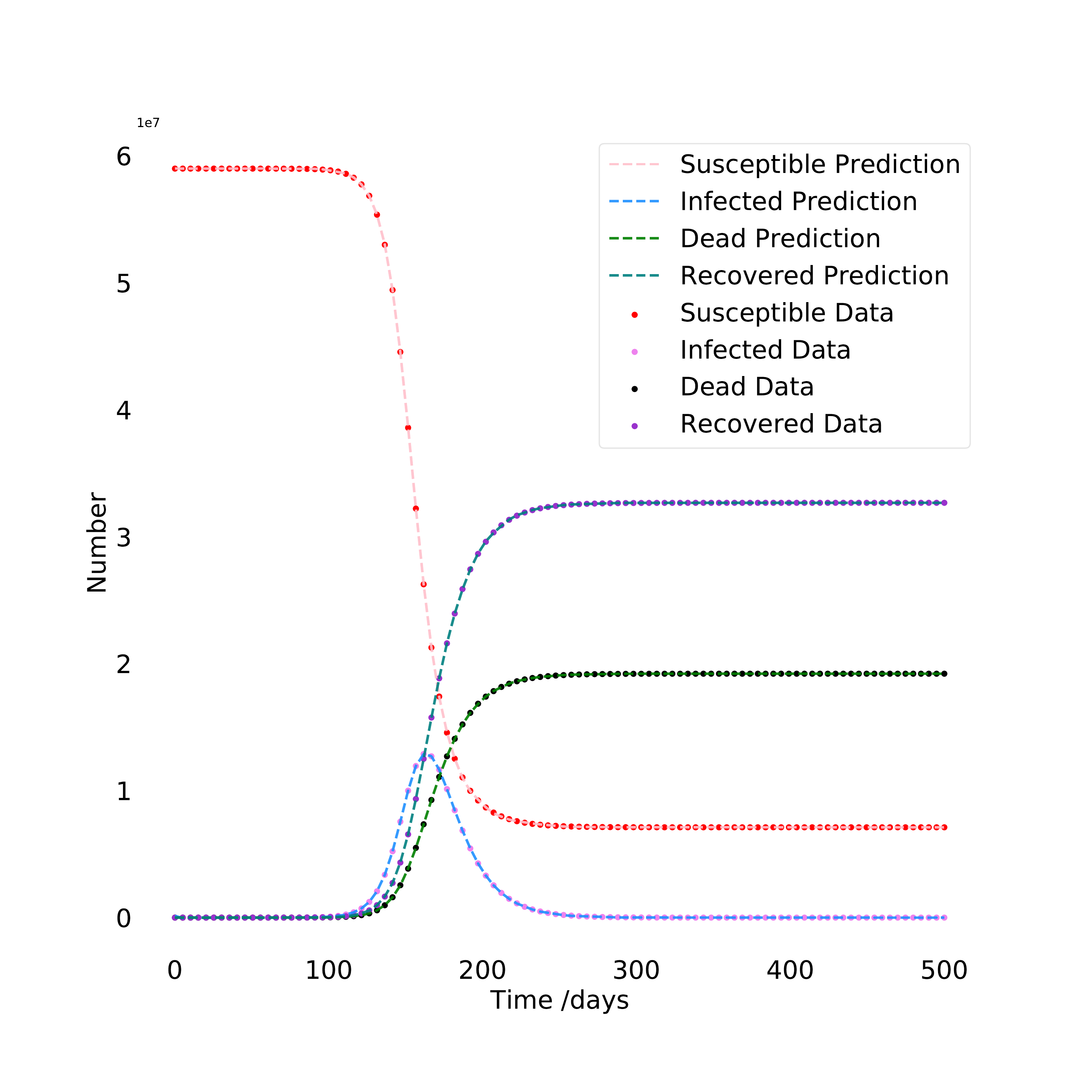}
    \caption{COVID: Neural Network Output}
    \label{fig:fig1}
\label{COVID_figure_appendix}
\end{figure}

The DINNs COVID system considered was given by
\begin{gather*}
\begin{align*}
&dS/dt = - (\alpha / N) S I&\\
&dI/dt = (\alpha / N) S I - \beta I - \gamma I& \\
&dD/dt = \gamma I&\\
&dR/dt = \beta I&
\end{align*}
\end{gather*}
The model used 8 layers with 20 neurons per layer, $1 \times 10^{-6}$ min learning rate, and was trained for 400k iterations (about 20 minutes). Figure \ref{COVID_figure_appendix} and table \ref{COVID_table2_appendix} show our results. The relative error corresponding to the SIDR system was $(0.022, 0.082, 0.022, 0.014)$.
\begin{table}[H]
\centering
\caption{COVID: Parameter Estimation}
\begin{tabular}[t]{lcccc}
\hline
Parameter & Actual Value & Range & Parameter Found & \% Relative Error\\
\hline
$\alpha$&0.191&(-1,1) & 0.1932& 1.151\\
$\beta$&0.05&(-1,1) & 0.0501& 0.2\\
$\gamma$&0.0294&(-1,1) & 0.0297& 1.02\\
\hline
\end{tabular}
\label{COVID_table2_appendix}
\end{table}%

\subsubsection{HIV}
\begin{figure}[t]%
    \centering
    \includegraphics[width=0.6\columnwidth]{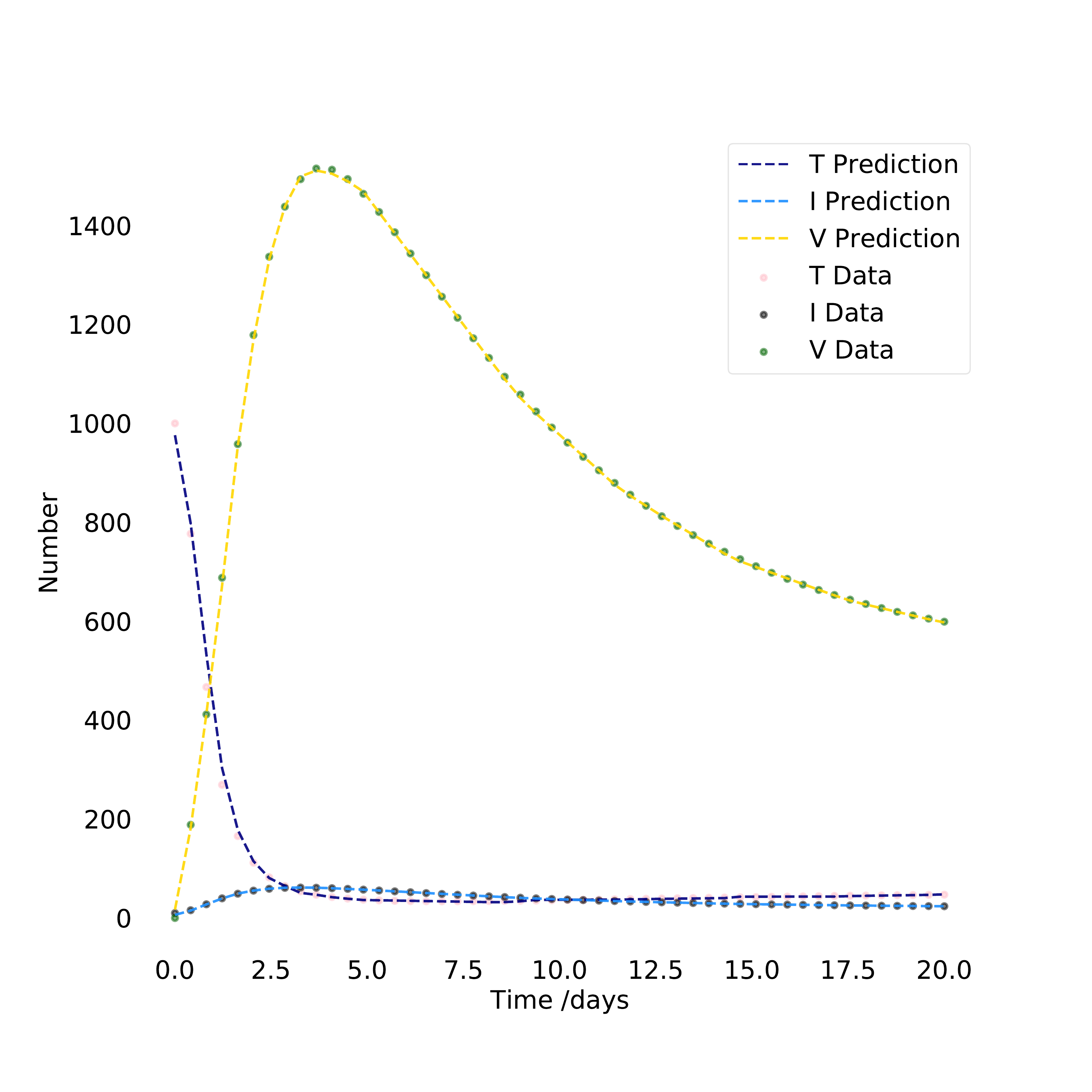}
    \caption{HIV: Neural Network Output}
    \label{fig:fig1}
\label{HIV_apdx_fig1}
\end{figure}

The DINN HIV model had 8 layers with 20 neurons per layer, $1\times 10^{-8}$ min learning rate, and was trained for 25mil iterations (about 22 hours). Figure \ref{HIV_apdx_fig1} and table \ref{HIV_apdx_table2} show our results. The relative error corresponding to the system was $(0.008, 0.002, 0.003)$.\\

System:
\begin{gather*}
\begin{align*}
&dT/dt = s - \mu_T  T + r  T  (1 - ((T + I) / T_{max}) - k_1  V  T)&\\
&dI/dt = k_1'  V  T - \mu_I  I& \\
&dV/dt = N  \mu_b  I - k_1  V  T - \mu_V  V&
\end{align*}
\end{gather*}

\begin{table}[H]
\centering
\caption{HIV: Parameter Estimation}
\begin{tabular}[t]{lcccc}
\hline
Parameter & Actual Value & Range & Parameter Found & \% Relative Error\\
\hline
$s$&10&(9.9,10.1) & 10.000751 & 0.007515 \\
$\mu_T$&0.02&(0.018,0.022) & 0.020762 & 3.811689\\
$\mu_I$&0.26&(0.255,0.265) & 0.261271 & 0.488758\\
$\mu_b$&0.24&(0.235,0.245) & 0.241747 & 0.727760\\
$\mu_V$&2.4&(2.5,2.3) & 2.419914 & 0.829756\\
$r$&0.03&(0.029,0.031) & 0.030605 & 2.015910\\
$N$&250&(247.5,252.5) & 249.703094 & 0.118762\\
$T_{max}$&1500&(1485,1515) & 1506.543823 & 0.436255\\
$k_1$&$2.4\cdot10e^{-5}$&$(2.3\cdot10e^{-5},2.6\cdot10e^{-5})$ & 0.000246 & 2.447948\\
$k_1'$&$2\cdot10e^{-5}$&$(1.9\cdot10e^{-5},2.1\cdot10e^{-5})$ & 0.000203 & 1.599052\\
\hline
\end{tabular}
\label{HIV_apdx_table2}
\end{table}%

\subsubsection{Smallpox}
\begin{figure}[t]%
    \centering
    \includegraphics[width=0.6\columnwidth]{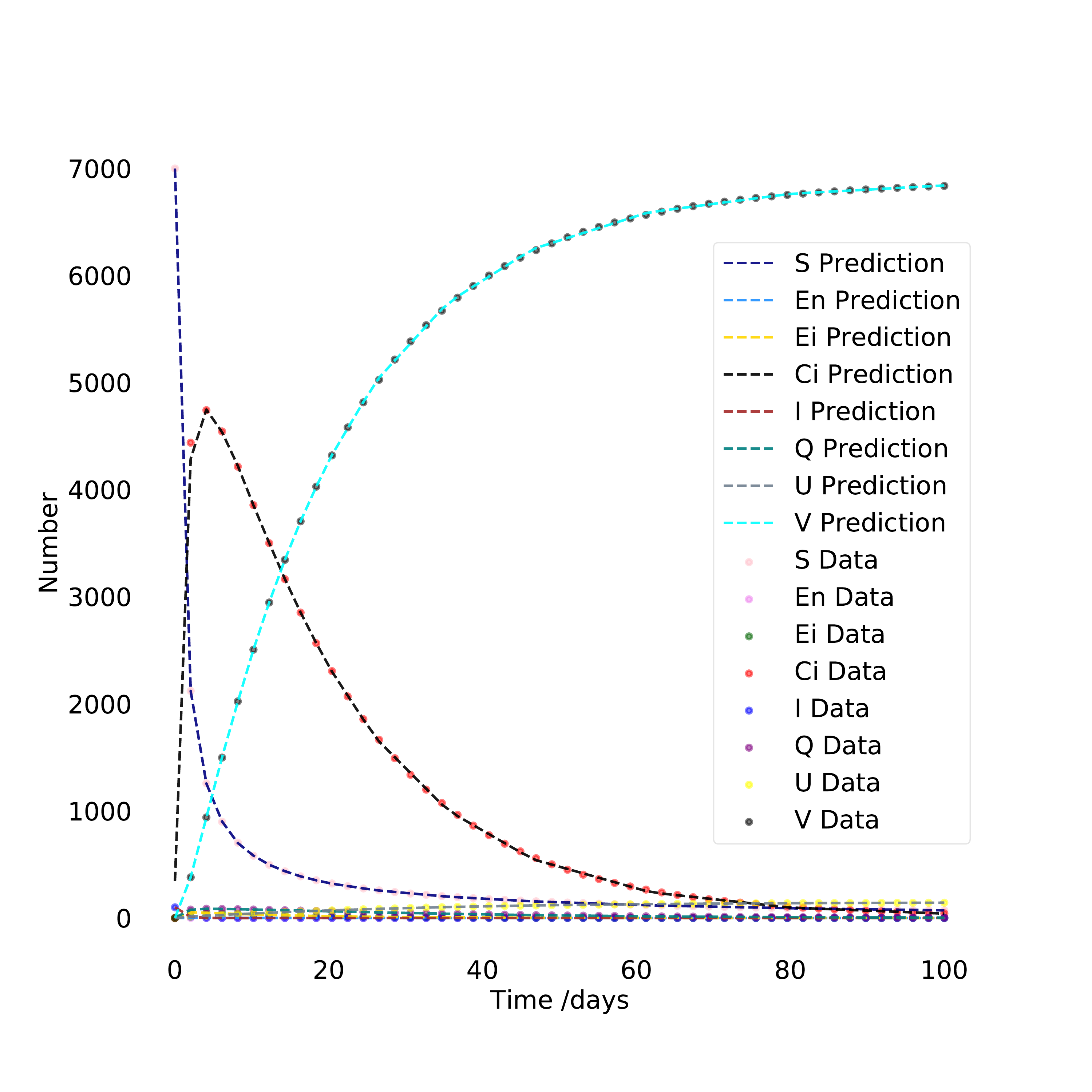}
    \caption{Smallpox: Neural Network Output}
    \label{fig:fig2}
\label{smallpox_apdx_fig}
\end{figure}

The DINN Smallpox model had 8 layers with 20 neurons per layer, $1e^{-7}$ min learning rate, and was trained for 12mil iterations (about 14 hours). Figure \ref{smallpox_apdx_fig} and table \ref{smallpox_apdx_table2} show our results. The relative error corresponding to the system was \\
$(0.033, 0.053, 0.045, 0.060, 0.014, 0.036, 0.027, 0.021)$. \\

System:
\begin{gather*}
\begin{align*}
&dS/dt = \chi_1  (1-\epsilon_1)  Ci - \beta  (\phi + \rho - \phi  \rho)  S  I&\\
&dEn/dt = \beta  \phi  (1-\rho)SI - \alpha En& \\
&dEi/dt = \beta  \phi\rho SI - (\chi_1\epsilon_2 + \alpha(1-\epsilon_2)) Ei&\\
&dCi/dt = \beta\rho(1-\phi)SI-\chi_1Ci&\\
&dI/dt = \alpha(1-\theta)En- (\theta+\gamma)I&\\
&dQ/dt = \alpha(1-\epsilon_2)Ei+\theta(\alpha En+I)-\chi_2Q&\\
&dU/dt = \gamma I+\chi_2Q&\\
&dV/dt = \chi_1(\epsilon_2Ei+\epsilon_1Ci) &
\end{align*}
\end{gather*}

\begin{table}[H]
\centering
\caption{Smallpox: Parameter Estimation}
\begin{tabular}[t]{lcccc}
\hline
Parameter & Actual Value & Range & Parameter Found & \% Relative Error\\
\hline
$\chi_1$&0.06&(0.054,0.066) & 0.0554 & 7.7222\\
$\chi_2$&0.04&(0.036,0.044) & 0.0380 & 4.9239\\
$\epsilon_1$&0.975&(0.86,1.04) & 0.9839 & 0.9089\\
$\epsilon_2$&0.3&(0.27,0.33) & 0.2841 & 5.2848\\
$\rho$&0.975&(0.86,1.04) & 0.9759 & 0.0882\\
$\theta$&0.95&(0.86,1.04) & 0.9050 & 4.7371\\
$\alpha$&0.068&(0.061,0.075) & 0.0626 & 8.5490\\
$\gamma$&0.11&(0.10,0.12) & 0.1034 & 10.8598\\
\hline
\end{tabular}
\label{smallpox_apdx_table2}
\end{table}%

\subsubsection{Tuberculosis}
\begin{figure}[t]%
    \centering
    \includegraphics[width=0.6\columnwidth]{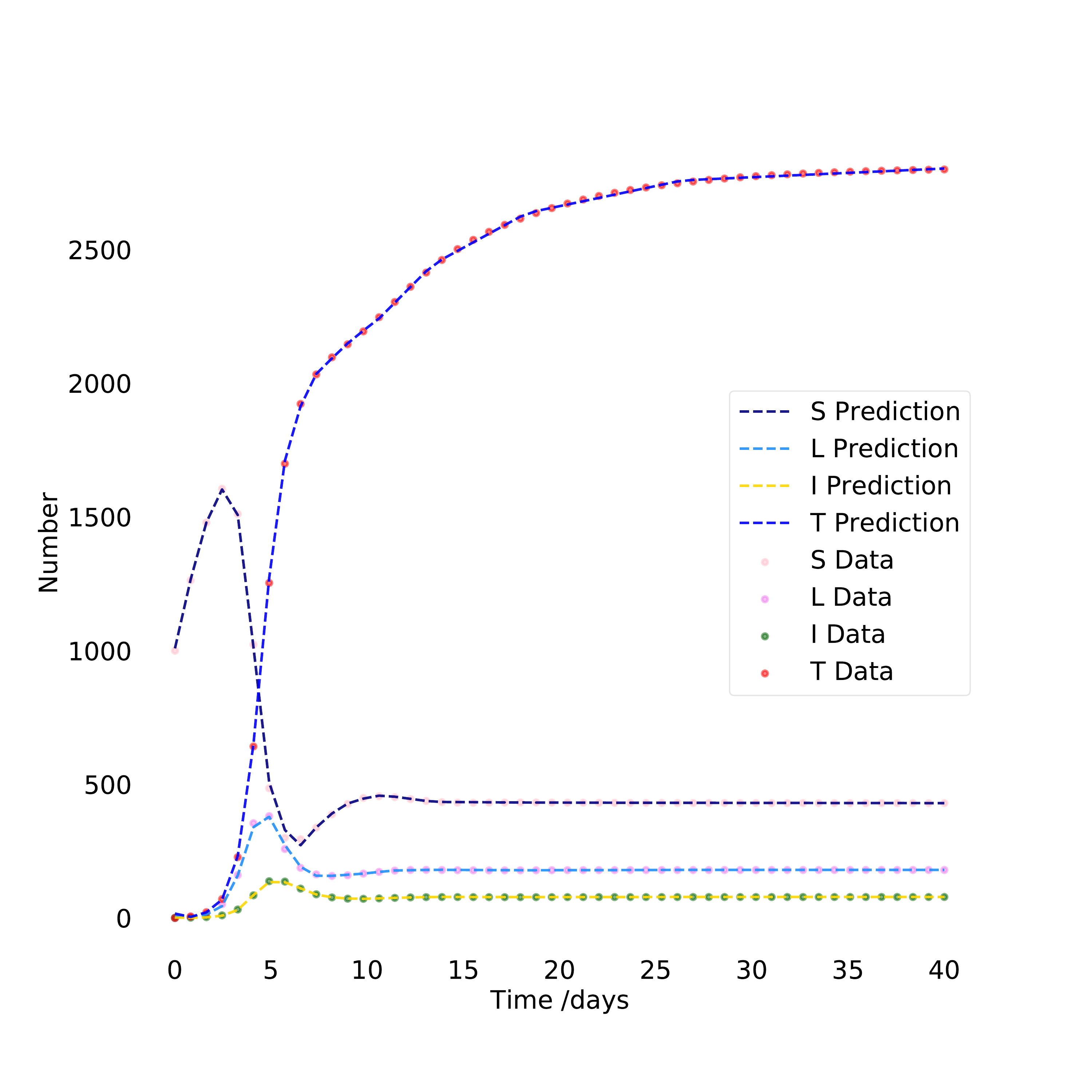}
    \caption{Tuberculosis: Neural Network Output}
    \label{fig:fig1}
\label{tb_apdx_fig}
\end{figure}

The DINN Tuberculosis model had 8 layers with 20 neurons per layer, $1e^{-7}$ min learning rate, and was trained for 10mil iterations (about 12 hours).  Figure \ref{tb_apdx_fig} and table \ref{tb_apdx_table2} show our results. The relative error corresponding to the system was $(0.030, 0.034, 0.034, 0.008)$.\\

System:
\begin{gather*}
\begin{align*}
&dS/dt = \delta - \beta  c  S  I / N - \mu  S&\\
&dL/dt = \beta  c  S  I / N - (\mu + k + r_1)  L +
\beta'  c  T /N& \\
&dI/dt = kL - (\mu + d)  I - r_2  I&\\
&dT/dt = r_1  L + r_2  I - \beta'  c  T /N - \mu T&
\end{align*}
\end{gather*}

\begin{table}[H]
\centering
\caption{Tuberculosis: Parameter Estimation}
\begin{tabular}[t]{lcccc}
\hline
Parameter & Actual Value & Range & Parameter Found & \% Relative Error\\
\hline
$\delta$&500&(480,520) & 509.4698 & 1.8587 \\
$\beta$&13&(9,15) & 12.5441 & 3.6341 \\
$c$&1&(-1,3) & 1.0405 & 3.8952 \\
$\mu$&0.143&(0.1,0.3) & 0.1474 & 3.0142 \\
$k$&0.5&(0,1) & 0.5396 & 7.3433 \\
$r_1$&2&(1,3) & 1.9892 & 0.5424 \\
$r_2$&1&(-1,3) & 1.1243 & 11.0583 \\
$\beta'$&13&(9,15) & 13.7384 & 5.3746 \\
$d$&0&(-0.4,0.4) & -0.0421 & 0.0421 \\
\hline
\end{tabular}
\label{tb_apdx_table2}
\end{table}%

\subsubsection{Pneumonia}
\begin{figure}[t]%
    \centering
    \includegraphics[width=0.7\columnwidth]{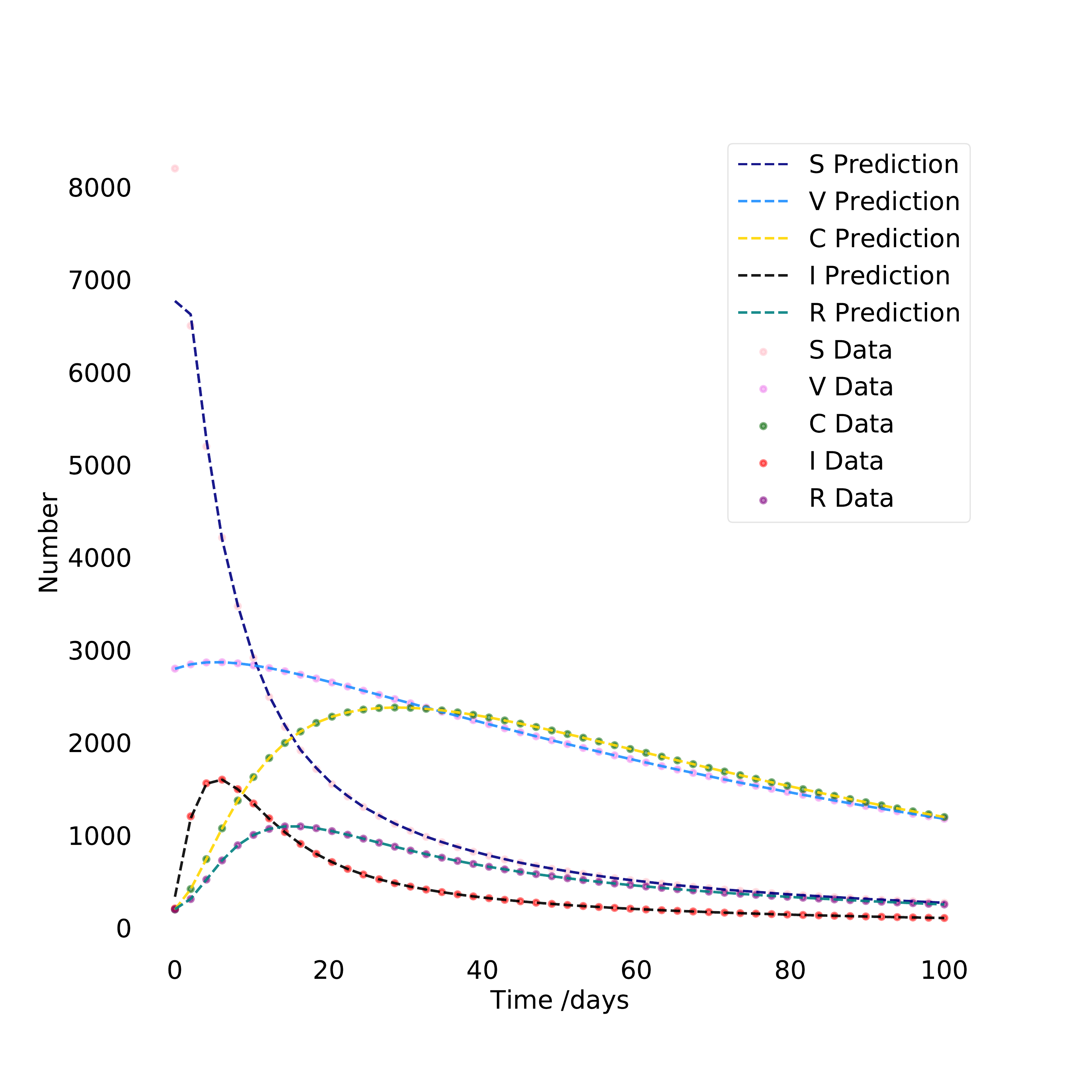}
    \caption{Pneumonia: Neural Network Output}
    \label{fig:fig3}
\label{pn_apdx_fig}
\end{figure}

The DINN Pneumonia model had 8 layers with 64 neurons per layer, $1\times 10^{-7}$ min learning rate, and was trained for 25mil iterations (about 41 hours). Figure \ref{pn_apdx_fig} and tables \ref{pn_apdx_table2} show our results.  The relative error corresponding to the system was $(0.020, 0.039, 0.034, 0.019, 0.023)$.\\

System:
\begin{gather*}
\begin{align*}
&dS/dt = (1-p)\pi + \phi V + \delta  R - (\mu + \lambda + \theta)  S&\\
&dV/dt = p  \pi + \theta  S - (\mu + \epsilon  \lambda + \phi)  V& \\
&dC/dt = \rho  \lambda  S + \rho \epsilon \lambda V + (1-q)  \eta  I - (\mu + \beta + \chi)  C&\\
&dI/dt = (1-\rho)\lambda  S + (1-\rho) \epsilon \lambda V+\chi C-(\mu+\alpha+\eta)I&\\
&dR/dt = \beta  C + q \eta I - (\mu + \delta)R&
\end{align*}
\end{gather*}

\begin{table}[H]
\centering
\caption{Pneumonia: Parameter Estimation}
\begin{tabular}[t]{lcccc}
\hline
Parameter & Actual Value & Range & Parameter Found & \% Relative Error\\
\hline
$\pi$&0.01&(0.0099,0.011) & 0.0098 & 2.0032\\
$\lambda$&0.1&(0.099,0.11) & 0.0990 & 0.9622\\
$k$&0.5&(0.49,0.51) & 0.5025 & 0.5083\\
$\epsilon$&0.002&(0.001,0.003) & 0.0022 & 11.6847\\
$\tau$&0.89&(0.87,0.91) & 0.8912 & 0.1309\\
$\phi$&0.0025&(0.0023,0.0027) & 0.0027 & 7.4859\\
$\chi$&0.001&(0.0009,0.0011) & 0.0011 & 6.7374\\
$p$&0.2&(0.19, 0.21) & 0.2033 & 1.6372\\
$\theta$&0.008&(0.0075,0.0085) & 0.0084 & 4.8891\\
$\mu$&0.01&(0.009,0.011) & 0.0092 & 8.4471\\
$\alpha$&0.057&(0.056,0.058) & 0.0570 & 0.0005\\
$\rho$&0.05&(0.049,0.051) & 0.0508 & 1.5242\\
$\beta$&0.0115&(0.0105,0.0125) & 0.0122 & 5.8243\\
$\eta$&0.2&(0.19,0.21) & 0.2023 & 1.1407\\
$q$&0.5&(0.49,0.51) & 0.4960 & 0.8003\\
$\delta$&0.1&(0.09,0.11) & 0.1038 & 3.7502\\
\hline
\end{tabular}
\label{pn_apdx_table2}
\end{table}%

\subsubsection{Ebola}
\begin{figure}[t]%
    \centering
    \includegraphics[width=0.6\columnwidth]{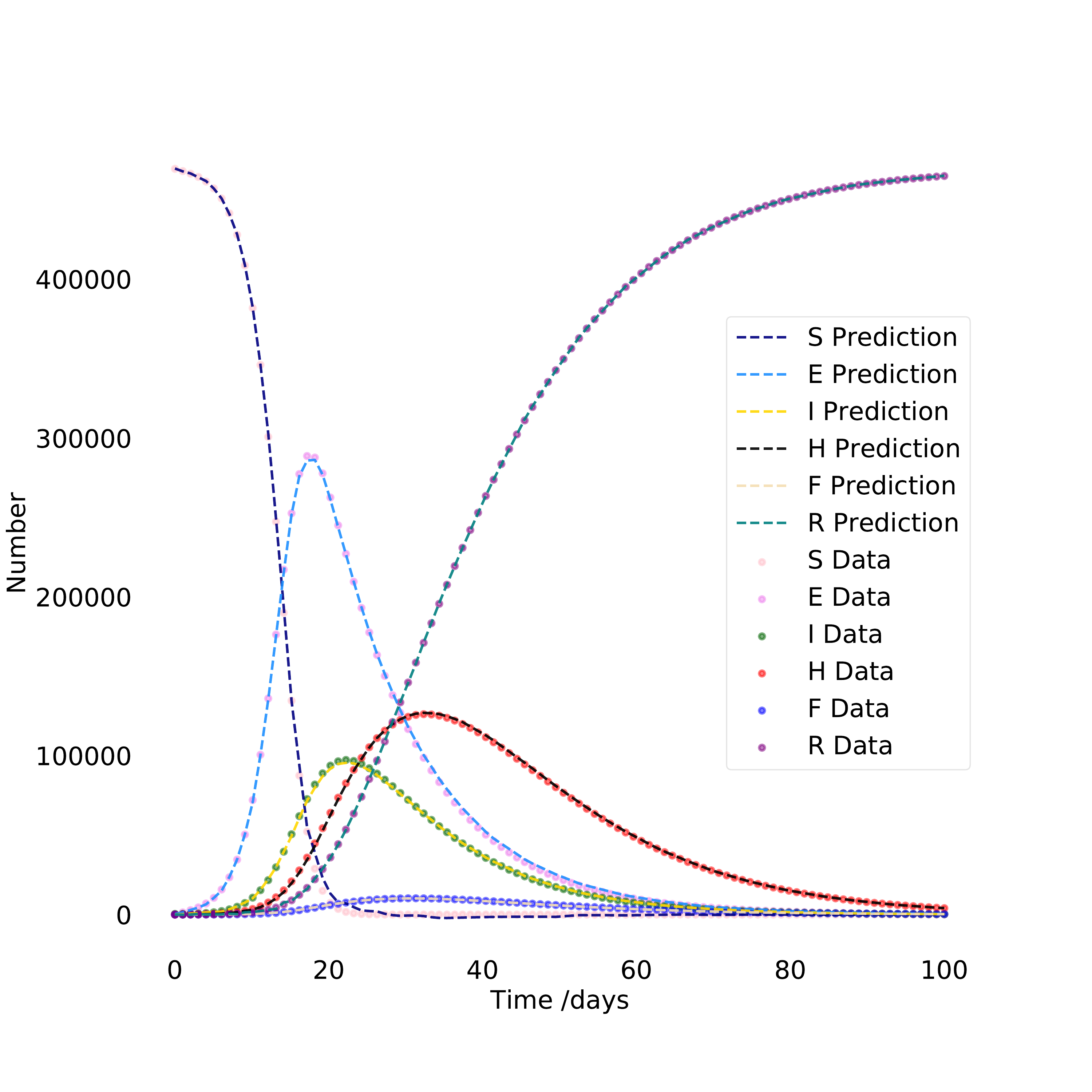}
    \caption{Ebola: Neural Network Output}
    \label{fig:fig3}
\label{eb_apdx_fig}
\end{figure}

Next, we consider the Ebola model considered before. The DINN Ebola model had 8 layers with 20 neurons per layer, $1e^{-7}$ min learning rate, and was trained for 20mil iterations (about 33 hours). Figure \ref{eb_apdx_fig} and table \ref{eb_apdx_table2} show our results. The relative error corresponding to the SIDR system was $(0.023, 0.050, 0.044, 0.062, 0.049, 0.005)$\\

\begin{table}[H]
\centering
\caption{Ebola: Parameter Estimation}
\begin{tabular}[t]{lcccc}
\hline
Parameter & Actual Value & Range & Parameter Found & \% Relative Error\\
\hline
$\beta_1$&3.532&(3.5,3.56) & 3.5589 & 0.7622 \\
$\beta_h$&0.012&(0.011,0.013) & 0.0129 & 7.8143 \\
$\beta_f$&0.462&(0.455,0.465) & 0.4638 & 0.3976 \\
$\alpha$&1/12&(0.072,0.088) & 0.0866 & 3.9320 \\
$\gamma_h$&1/4.2&(0.22,0.28) & 0.2471 & 3.7853 \\
$\theta_1$&0.65&(0.643,0.657) & 0.6523 & 0.3477 \\
$\gamma_i$&0.1&(0.099,0.11) & 0.0904 & 9.6403 \\
$\delta_1$&0.47&(0.465,0.475) & 0.4712 & 0.2565 \\
$\gamma_d$&1/8&(0.118,0.122) & 0.1205 & 3.6124 \\
$\delta_2$&0.42&(0.415,0.425) & 0.4247 & 1.1244 \\
$\gamma_f$&0.5&(0.45,0.55) & 0.5196 & 3.9246 \\
$\gamma_{ih}$&0.082&(0.081,0.083) & 0.0811 & 1.0932 \\
$\gamma_{dh}$&0.07&(0.069,0.071) & 0.0710 & 0.7563 \\
\hline
\end{tabular}
\label{eb_apdx_table2}
\end{table}%
\begin{figure}[h!]%
    \centering
    \includegraphics[width=0.6\columnwidth]{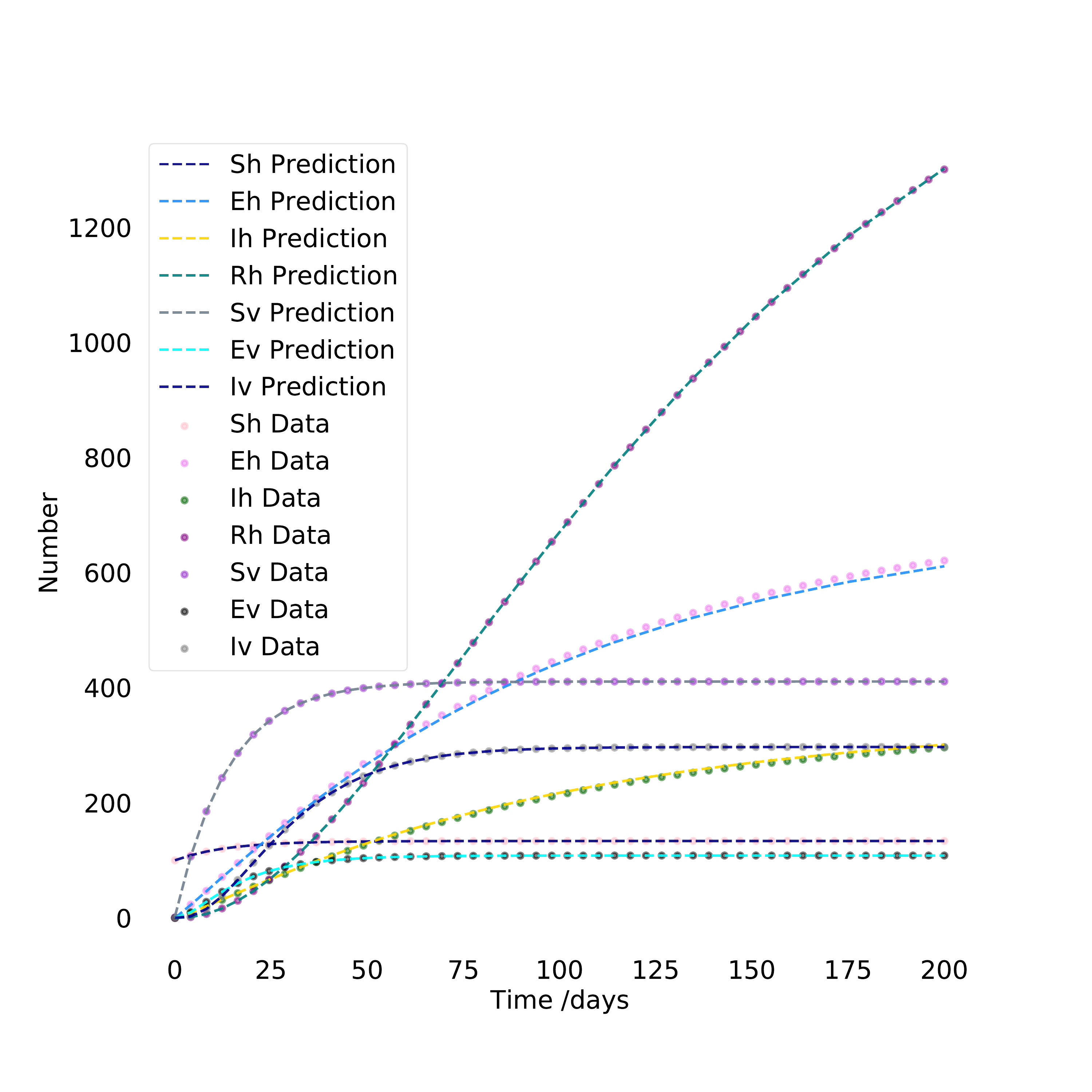}
    \caption{Dengue: Neural Network Output}
    \label{fig:fig3}
\label{dg_apdx_fig}
\end{figure}
\subsubsection{Dengue}
The DINN Dengue model had 8 layers with 20 neurons per layer, $1e^{-7}$ min learning rate, and was trained for 35mil iterations (about 58 hours). Figure \ref{dg_apdx_fig} and table  \ref{dg_apdx_table2} show our results.The relative error is $(0.003, 0.012, 0.030, 0.054, 0.001, 0.001, 0.002)$.\\

System:
\begin{eqnarray*}
dSh/dt &=& \pi_h - \lambda_h  Sh - \mu_h  Sh\\
dEh/dt &=& \lambda_h  Sh - (\sigma_h  \mu_h)  Eh \\
dIh/dt &=& \sigma_h  Eh - (\tau_h + \mu_h + \delta_h)  Ih\\
dRh/dt &=& \tau_h Ih - \mu_h Rh\\
dSv/dt &=& \pi_v - \delta_v  Sv - \mu_v  Sv\\
dEv/dt &=& \delta_v  Sv - (\sigma_v + \mu_v)  Ev\\
dIv/dt &=& \sigma_v  Ev - (\mu_v + \delta_v)  Iv
\end{eqnarray*}

\begin{table}[H]
\centering
\caption{Dengue: parameters, their values, the parameters search range that DINN was trained on, the parameters found after training, and the relative error percentage}
\begin{tabular}[t]{lcccc}
\hline
Parameter & Actual Value & Range & Parameter Found & \% Relative Error\\
\hline
$\pi_h$&10&(9.9,10.1) & 9.9317 & 0.6832\\
$\pi_v$&30&(29.7,30.3) & 29.8542 & 0.4859\\
$\lambda_h$&0.055&(0.054,0.056) & 0.0552 & 0.2696\\
$\lambda_v$&0.05&(0.049,0.051) & 0.0506 & 1.2876\\
$\delta_h$&0.99&(0.9,1.1) & 0.9643 & 2.5967\\
$\delta_v$&0.057&(0.056,0.058) & 0.0567 & 0.5294\\
$\mu_h$&0.0195&(0.0194,0.0196) & 0.0194 & 0.3835\\
$\mu_v$&0.016&(0.015,0.017) & 0.0159 & 0.8796\\
$\sigma_h$&0.53&(0.52,0.54) & 0.5372 & 1.3567\\
$\sigma_v$&0.2&(0.19,0.21) & 0.1989 & 0.5483\\
$\tau_h$&0.1&(0.05,0.15) & 0.0902 & 9.7723\\
\hline
\end{tabular}
\label{dg_apdx_table2}
\end{table}%

\subsubsection{Anthrax}
\begin{figure}[t]%
    \centering
    \includegraphics[width=0.6\columnwidth]{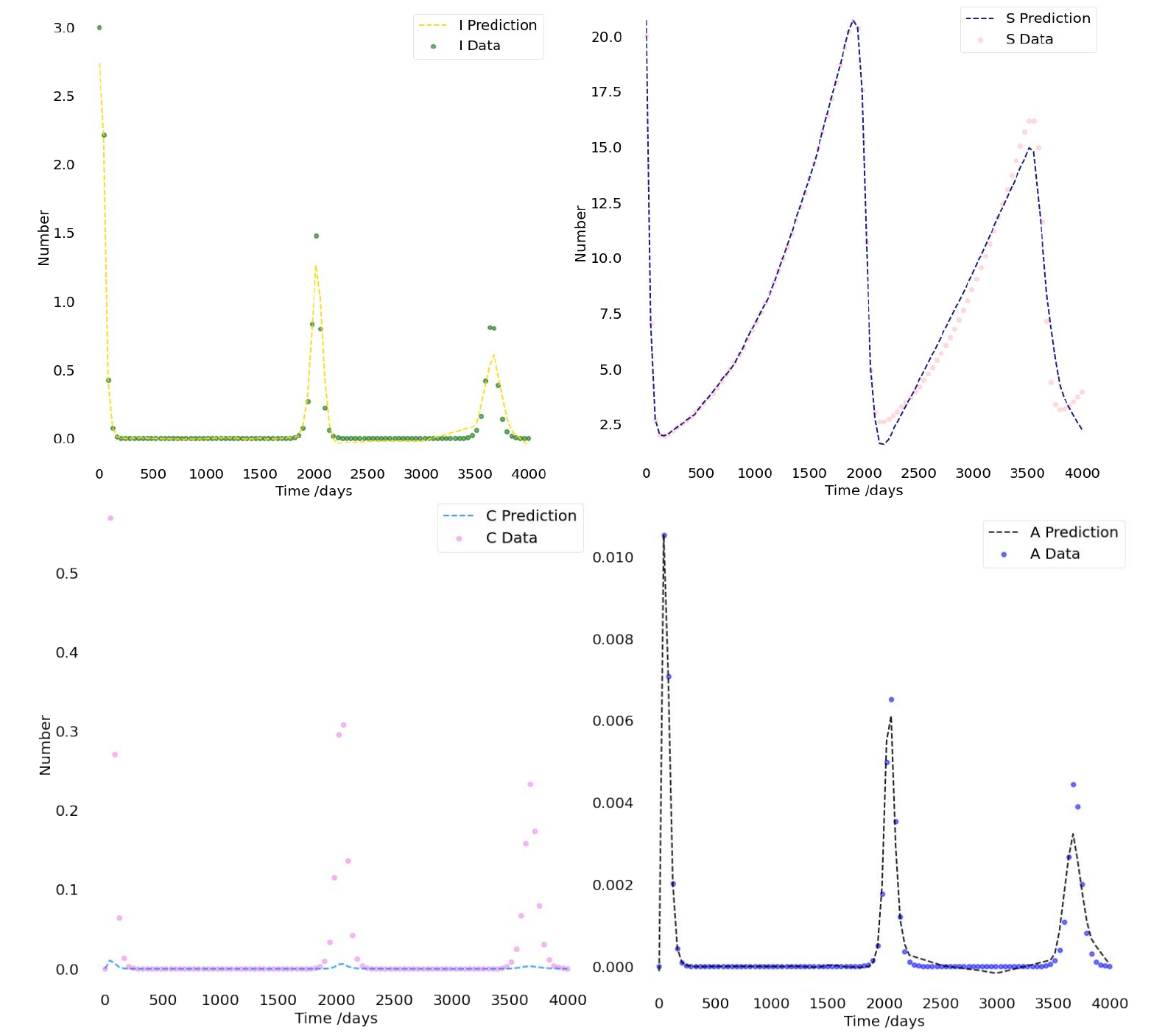}
    \caption{Anthrax: Neural Network Output}
    \label{fig:fig3}
\label{atx_apdx_fig}
\end{figure}

The DINN Anthrax model had 8 layers with 64 neurons per layer, $1e^{-8}$ min learning rate, and was trained for 55mil iterations (about 91 hours). Figure \ref{atx_apdx_fig} and tables \ref{atx_apdx_table1}, \ref{atx_apdx_table2} show our results.\\

System:
\begin{gather*}
\begin{align*}
&dS/dt = r  (S + I)  (1 - (S + I)/K) - \eta_a  A  S - \eta_c  S  C - \eta_i  (S  I)/(S + I) - \mu  S + \tau  I&\\
&dI/dt = \eta_a  A  S + \eta_c  S  C + (\eta_i  (S  I)/(S + I) - (\gamma + \mu + \tau))  I& \\
&dA/dt = -\sigma  A + \beta  C&\\
&dC/dt = (\gamma + \mu)  I - \delta  (S + I)  C - \kappa  C&
\end{align*}
\end{gather*}

\begin{table}[H]
\centering
\caption{Anthrax: relative error from LSODA generation of the learnable parameters}
\begin{tabular}{lS[table-format=1.4]cS[table-format=1.4]}
\toprule
\thead{ \thead{(S, I, A, C) Error} } \\
\midrule
(0.052, 0.144, 0.171, 0.171) \\
\bottomrule
\end{tabular}
\label{atx_apdx_table1}
\end{table}

\begin{table}[H]
\centering
\caption{Anthrax: parameters, their values, the parameters search range that DINN was trained on, the parameters found after training, and the relative error percentage}
\begin{tabular}[t]{lcccc}
\hline
Parameter & Actual Value & Range & Parameter Found & \% Relative Error\\
\hline
r&1/300&(0.003,0.0036) & 0.0034 & 1.2043\\
$\mu$&1/600&(0.0014,0.0018) & 0.0017 & 0.5754\\
$\kappa$&0.1&(0.99,0.11) & 0.1025 & 2.5423\\
$\eta_a$&0.5&(0.49,0.51) & 0.5035 & 0.7022\\
$\eta_c$&0.1&(0.09,0.11) & 0.1024 & 2.4492\\
$\eta_i$&0.01&(0.09,0.011) & 0.0106 & 6.0459\\
$\tau$&0.1&(0.09,0.11) & 0.0976 & 2.4492\\
$\gamma$&1/7&(0.13,0.15) & 0.1444 & 1.0542\\
$\delta$&1/64&(0.03,0.07) & 0.0512 & 2.3508\\
$K$&100&(98,102) & 100.6391 & 0.6391\\
$\beta$&0.02&(0.0018,0.0022) & 0.0021 & 6.5466\\
$\sigma$&0.1&(0.09,0.11) & 0.1051 & 5.1029\\
\hline
\end{tabular}
\label{atx_apdx_table2}
\end{table}%

\subsubsection{Polio}
\begin{figure}[t]%
    \centering
    \includegraphics[width=0.6\columnwidth]{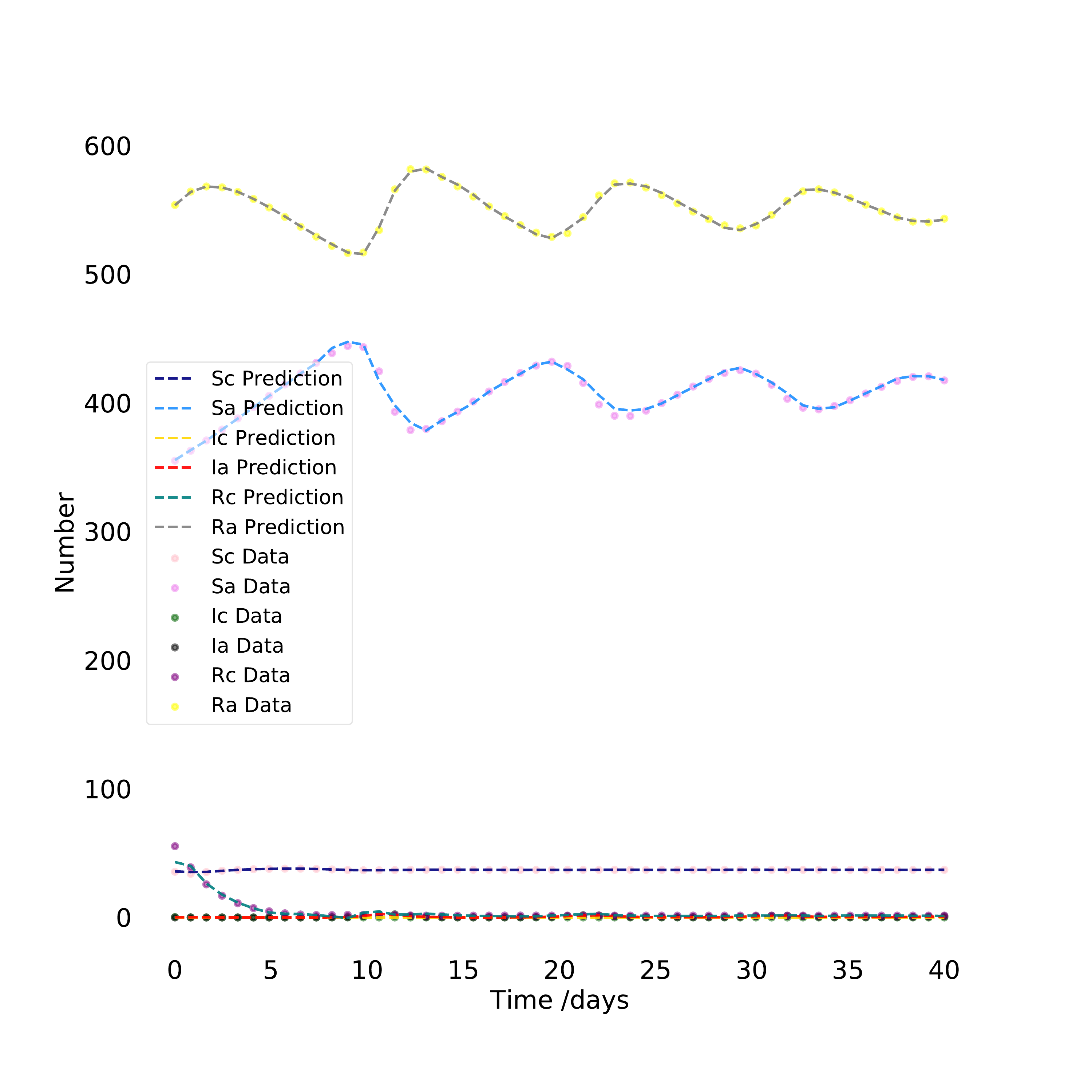}
    \caption{Polio: Neural Network Output}
\label{pl_apdx_fig}
\end{figure}

The DINN Polio model had 8 layers with 64 neurons per layer, $1e^{-8}$ min learning rate, and was trained for 40mil iterations (about 66 hours). Figure \ref{pl_apdx_fig} and tables \ref{pl_apdx_table1}, \ref{pl_apdx_table2} show our results.\\

System:
\begin{gather*}
\begin{align*}
&dSc/dt = \mu N - (\alpha+\mu+(\beta_{cc}/Nc)  Ic + (\beta_{ca}/Nc)  Ia)   Sc&\\
&dSa/dt = \alpha Sc - (\mu + (\beta_{aa}/Na) Ia + (\beta_{ac}/Na) Ic) Sa& \\
&dIc/dt = ((\beta_{cc}/Nc) Ic + (\beta_{ca}/Nc) Ia) Sc - (\gamma_c+\alpha+\mu) Ic&\\
&dIa/dt = ((\beta_{ac}/Na) Ic + (\beta_{aa}/Na) Ia) Sa - (\gamma_a+\mu) Ia +\alpha Ic&\\
&dRc/dt = \gamma_c Ic - \mu Rc - \alpha Rc&\\
&dRa/dt = \gamma_a Ia - \mu Ra + \alpha Rc&
\end{align*}
\end{gather*}

\begin{table}[H]
\centering
\caption{Polio: relative error from LSODA generation of the learnable parameters}
\begin{tabular}{lS[table-format=1.4]cS[table-format=1.4]}
\toprule
\thead{ \thead{(Sc, Sa, Ic, Ia, Rc, Ra) Error} } \\
\midrule
(0.001, 0.001, 0.017, 0.021, 0.004, 0.001) \\
\bottomrule
\end{tabular}
\label{pl_apdx_table1}
\end{table}

\begin{table}[H]
\centering
\caption{Polio: parameters, their values, the parameters search range that DINN was trained on, the parameters found after training, and the relative error percentage}
\begin{tabular}[t]{lcccc}
\hline
Parameter & Actual Value & Range & Parameter Found & \% Relative Error\\
\hline
$\mu$&0.02&(0.018,0.022) & 0.0200 & 0.0200\\
$\alpha$&0.5&(0.495,0.505) & 0.5018 & 0.36\\
$\gamma_a$&18&(17.9,18.1) & 18.0246 & 0.4168\\
$\gamma_c$&36&(35.8,36.2) & 36.0701 & 0.3587\\
$\beta_{aa}$&40&(39,41) & 40.2510 & 0.6275\\
$\beta_{cc}$&90&(89,91) & 90.6050 & 0.6722\\
$\beta_{ac}$&0&(-0.001,0.001) & 0.0002 & 0.0002\\
$\beta_{ca}$&0&(-0.001,0.001) & 0.0004 & 0.0004\\
\hline
\end{tabular}
\label{pl_apdx_table2}
\end{table}%

\subsubsection{Measles}
\begin{figure}[t]%
    \centering
    \includegraphics[width=0.6\columnwidth]{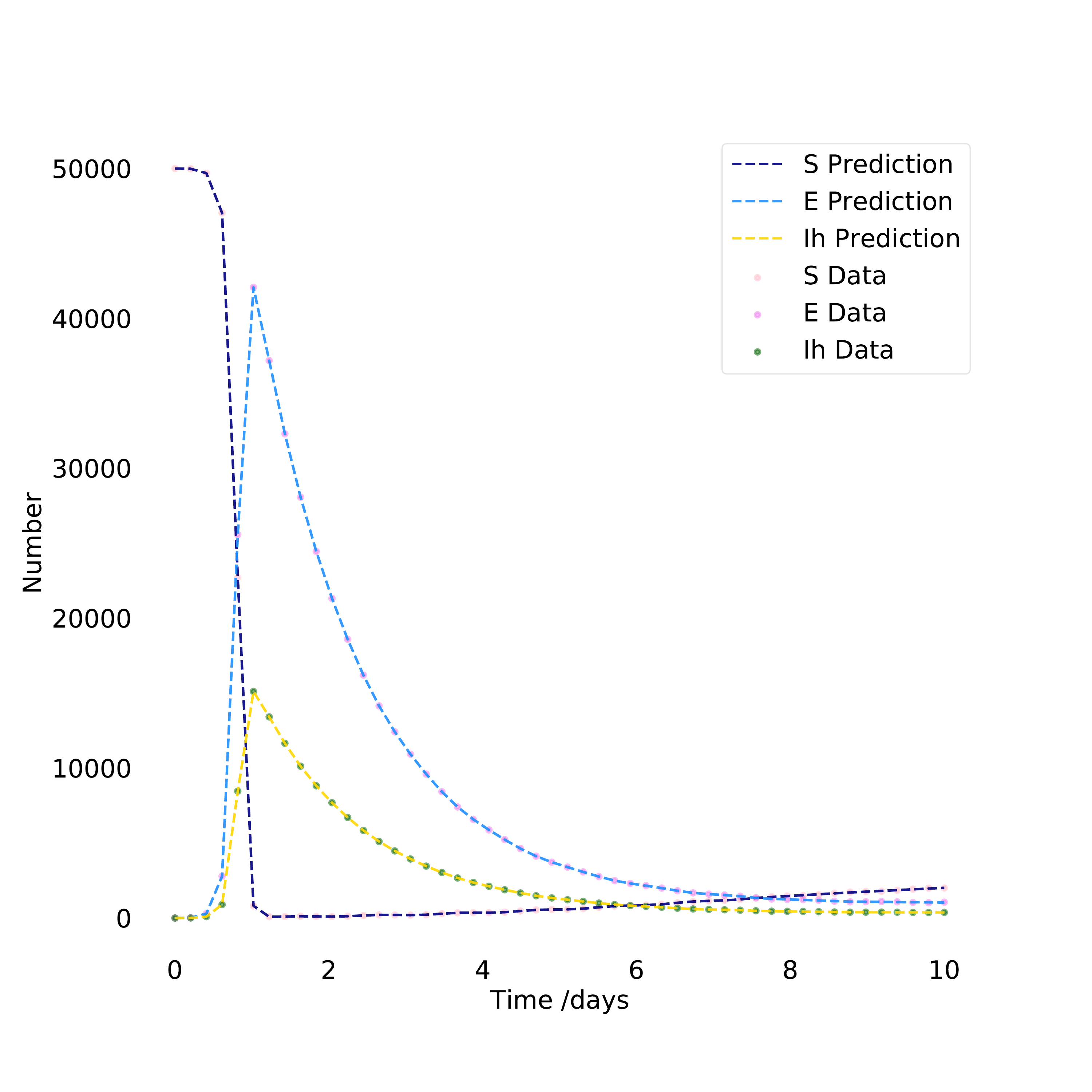}
    \caption{Measles: Neural Network Output}
\label{ms_apdx_fig}
\end{figure}

The DINN Measles model had 8 layers with 64 neurons per layer, $1e^{-7}$ min learning rate, and was trained for 17mil iterations (about 28 hours). Figure \ref{ms_apdx_fig} and tables \ref{ms_apdx_table1}, \ref{ms_apdx_table2} show our results.\\

System:
\begin{gather*}
\begin{align*}
&dS/dt = \mu  (N - S) - (\beta  S  I)/N&\\
&dE/dt = (\beta  S  I)/N - (\mu  \sigma)  E& \\
&dI/dt = \sigma  E - (\mu + \gamma)  I&
\end{align*}
\end{gather*}

\begin{table}[H]
\centering
\caption{Measles: relative error from LSODA generation of the learnable parameters}
\begin{tabular}{lS[table-format=1.4]cS[table-format=1.4]}
\toprule
\thead{ \thead{(S, E, I) Error} } \\
\midrule
(0.017, 0.058, 0.059) \\
\bottomrule
\end{tabular}
\label{ms_apdx_table1}
\end{table}

\begin{table}[H]
\centering
\caption{Measles: parameters, their values, the parameters search range that DINN was trained on, the parameters found after training, and the relative error percentage}
\begin{tabular}[t]{lcccc}
\hline
Parameter & Actual Value & Range & Parameter Found & \% Relative Error\\
\hline
$\mu$&0.02&(0.01,0.03) & 0.0225 & 12.704 \\
$\beta_1$&0.28&(0.27,0.37) & 0.2700 & 3.5704 \\
$\gamma$&100&(97,103) & 97.0001 & 2.9999 \\
$\sigma$&35.84&(33,37) & 34.7127 & 3.1453 \\
\hline
\end{tabular}
\label{ms_apdx_table2}
\end{table}%

\subsubsection{Zika}
\begin{figure}[t]%
    \centering
    \includegraphics[width=0.6\columnwidth]{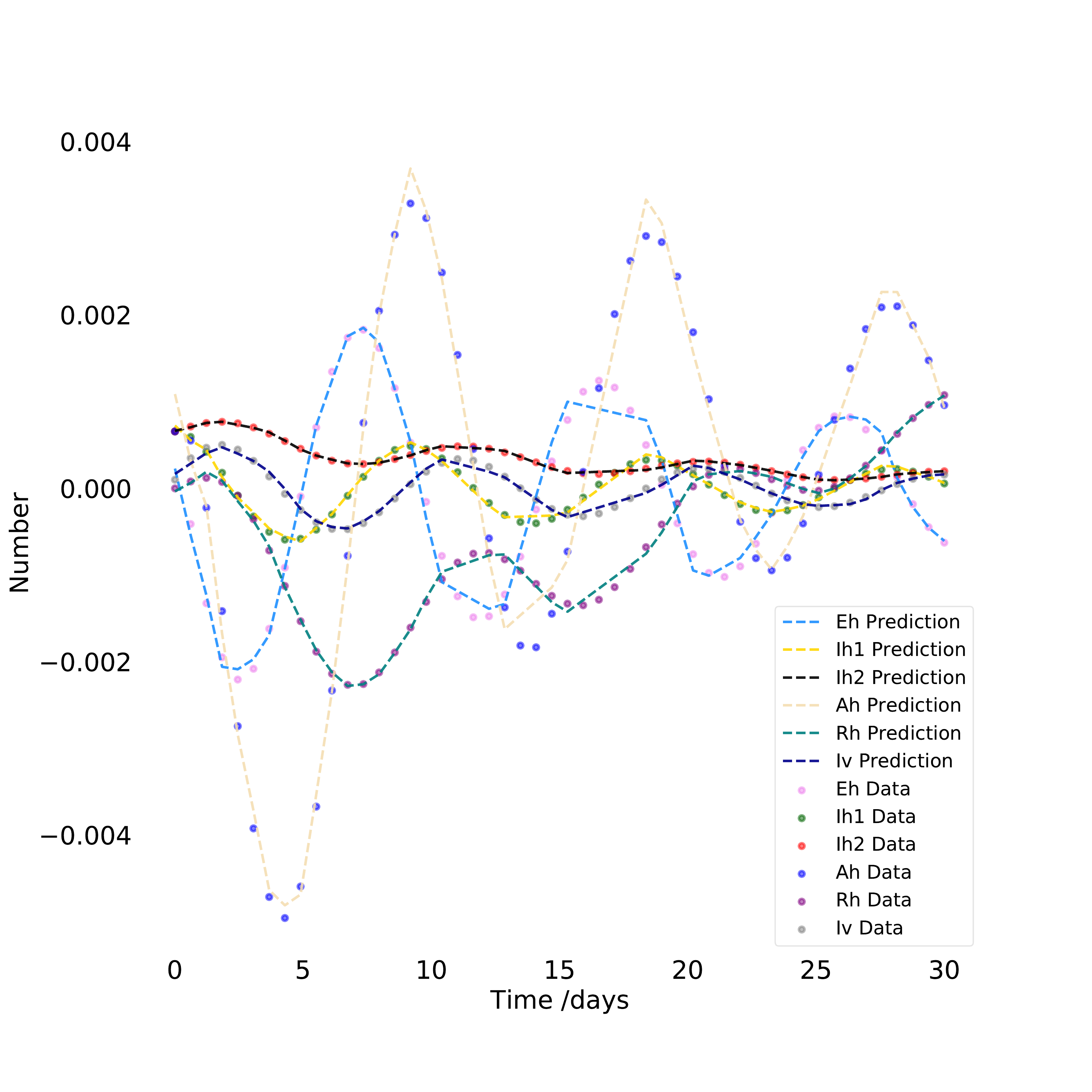}
    \caption{Zika: Neural Network Output}
\label{zk_apdx_fig}
\end{figure}

The DINN Zika model had 8 layers with 64 neurons per layer, $1e^{-9}$ min learning rate, and was trained for 8mil iterations (about 13 hours). The following image has a selection of the compartments to reduce scatter in visualization. Figure \ref{zk_apdx_fig} and tables \ref{zk_apdx_table1}, \ref{zk_apdx_table2} show our results.\\

System:
\begin{gather*}
\begin{align*}
&dSh/dt = -a  b  (Iv/Nh)  Sh - \beta  ((\kappa  Eh + Ih_1 + \tau  Ih_2) / Nh )  Sh&\\
&dEh/dt = \theta  (-a  b  (Iv/Nh)  Sh - \beta  ((\kappa  Eh + Ih_1 + \tau  Ih_2) / Nh )  Sh) - V_h  Eh& \\
&dIh_1/dt = V_h  Eh - \gamma_{h1}  Ih_1&\\
&dIh_2/dt = \gamma_{h1}  Ih_1 - \gamma_{h2}  Ih_2&\\
&dAh/dt = (1 - \theta)  (a  b  (Iv/Nh)  Sh - \beta  ((\kappa  Eh + Ih_1 + \tau
Ih_2) / Nh )  Sh) - \gamma_h  Ah& \\
&dRh/dt = \gamma_{h2}  Ih_2 + \gamma_h  Ah&\\
&dSv/dt = \mu_v  Nv - a  c  ((\eta  Eh + Ih_1)/Nh)  Sv - \mu_v  Sv&\\
&dEv/dt = a  c  ((\eta  Eh + Ih_1)/Nh) - (V_v + \mu_v)  Ev& \\
&dIv/dt = V_v  Ev - \mu_v  Iv&
\end{align*}
\end{gather*}

\begin{table}[H]
\centering
\caption{Zika: relative error from LSODA generation of the learnable parameters}
\begin{tabular}{lS[table-format=1.4]cS[table-format=1.4]}
\toprule
\thead{ \thead{($Sh, Ih_1, Ih_2, Ah, Rh, Sv, Ev, Iv, I$) Error} } \\
\midrule
($2.215e^{-06}$, 0.017, 0.014, 0.003, 0.024, 0.091, 0.005, 0.012, 0.018, 0.018) \\
\bottomrule
\end{tabular}
\label{zk_apdx_table1}
\end{table}

\begin{table}[H]
\centering
\caption{Zika: parameters, their values, the parameters search range that DINN was trained on, the parameters found after training, and the relative error percentage}
\begin{tabular}[t]{lcccc}
\hline
Parameter & Actual Value & Range & Parameter Found & \% Relative Error\\
\hline
a&0.5&(0.49,0.51) & 0.4997 & 0.0588 \\
b&0.4&(0.39,0.41) & 0.4033 & 0.8297 \\
c&0.5&(0.49,0.51) & 0.5015 & 0.3086 \\
$\eta$&0.1&(0.09,0.11) & 0.0999 & 0.0687 \\
$\beta$&0.05&(0.0495,0.0505) & 0.0498 & 0.4098 \\
$\kappa$&0.6&(0.594,0.606) & 0.6033 & 0.5486 \\
$\tau$&0.3&(0.27,0.33) & 0.2902 & 3.2565 \\
$\theta$&18&(0.17.8,18.2) & 17.9669 & 0.1838 \\
m&5&(4.5,5.5) & 5.2937 & 5.8748 \\
$V_h$&1/5&(0.198,0.202) & 0.1996 & 0.1798 \\
$V_v$&10&(9.9,10.1) & 10.0170 & 0.1700 \\
$\gamma_{h1}$&1/5&(0.18,0.22) & 0.1991 & 0.4651 \\
$\gamma_{h2}$&1/64&(0.045,0.055) & 0.0504 & 0.7261 \\
$\gamma_{h}$&1/7&(0.139,0.141) & 0.1406 & 1.5967 \\
$\mu_v$&1/14&(0.063,0.077) & 0.0723 & 1.1806 \\
\hline
\end{tabular}
\label{zk_apdx_table2}
\end{table}%

\section{Discussion and Conclusion}
\label{sec:conclusion}
In this work, we have introduced Disease Informed Neural Networks (DINNs) which is a neural network approach capable of learning a number of diseases, how they spread, forecasting their progression, and finding unique parameters that are used in models to describe the disease dynamics. Specifically, for a benchmark problem we were able to study the influence in ranges of parameter estimation, noise, data variability, NN architechture, learning rates and missing data on the performance of DINNs. Our results from this work suggest that DINNs is a robust and reliable candidate that can be used as an inverse approach to characterize and learn parameters used in compartmental models for understanding dynamics of infectious diseases. To compare the performance of the proposed DINNs, we also wrote the parameter estimation in R and MATLAB that employed powerful non-linear optimization methods such as Nelder-Mead, Gauss Newton and gradient decent methods. In all the types of simulations, we noticed DINNs outperformed and was more robust to initial parameter guesses. Especially, all the other methods failed to achieve the optimal solution if the initial guesses were far from the actual values for these optimization based methods compared to DINNs which worked extremely well.

Building on a simple SIRD model for COVID-19, we used it to model eleven infectious diseases and show the simplicity, efficacy, and generalization of DINNs in their respective applications. These diseases were modeled into various differential equations systems with various number of learnable parameters. We found that DINNs can quite easily learn systems with a low number of parameters and dimensions (e.g., COVID), and when the learnable parameters are known the training time can change from 50 hours to a few minutes. Moreover, it appears as if the number of dimensions does not affect the performance as much as the number of learnable parameters (e.g., see pneumonia vs ebola). From the anthrax model result we see that it is far more difficult for DINNs to learn systems which have numerous quick and sharp oscillations. That being said, looking at the polio and zika models results we can see that it is not impossible, but rather more time consuming (both in training and hyperparameter search). Also, based on the measles, tuberculosis, and smallpox models results we can see that a low number of sharp oscillations are relatively easy to learn. 

It maybe noted that while the goal of this work was to introduce a powerful algorithm for predicting infectious diseases, the algorithms have the potential to be applied to complex models (for example, involving spatial dependencies, using facemasks, impact of lockdowns, etc.). Also, while DINNs presented here is shown to be robust and reliable, it can be slow to train on particular problems and there is no known theoretical guarantee of corresponding error bounds. These will be explored in forthcoming papers.

\section{Acknowledgements}
\label{sec:Acknowledgements}
This work has been supported in part by the National Science Foundation DMS 2031027 and DMS 2031029.

\printbibliography






\end{document}